%% file: main.tex
\documentclass[lettersize,journal]{IEEEtran}

\input{misc/__misc__}
\begin{document}
\bstctlcite{IEEEexample:BSTcontrol}

\title{Dynamic Visual Semantic Sub-Embeddings and Fast Re-Ranking}

% \author{IEEE Publication Technology,~\IEEEmembership{Staff,~IEEE,}
\author{Wenzhang Wei, Zhipeng Gui, ~\IEEEmembership{Member, IEEE,} Changguang Wu, Dehua Peng, Anqi Zhao, Huayi Wu% <-this % stops a space
% \thanks{This paper was produced by the IEEE Publication Technology Group. They are in Piscataway, NJ.}% <-this % stops a space
\thanks{This work is supported by National Natural Science Foundation of China (No. 42090010, No. 41971349, No. 41930107) and National Key R\&D Plan of China “Intergovernmental International Scientific and Technological Innovation Cooperation” (No. 2021YFE0117000). (Corresponding author: Zhipeng Gui.)}% <-this % stops a space
\thanks{Wenzhang Wei, Zhipeng Gui are with School of Remote Sensing and Information Engineering, Wuhan University, Wuhan 430079, China (email: \{2022282130122, zhipeng.gui\}@whu.edu.cn).}% <-this % stops a space
\thanks{Changguang Wu is with the School of Computer Science and Engineering, Nanjing University of Science and Technology, Nanjing 210094, China (email: changguangwu@njust.edu.cn).}% <-this % stops a space
\thanks{Zhipeng Gui, Anqi Zhao, Dehua Peng, Huayi Wu are with State Key Laboratory of Information Engineering in Surveying, Mapping and Remote Sensing, Wuhan University, Wuhan 430079, China, also with Collaborative Innovation Center of Geospatial Technology, Wuhan University, Wuhan 430079, China (email: \{aqzhao, pengdh, wuhuayi\}@whu.edu.cn).}}

% The paper headers
\markboth{}%
{Shell \MakeLowercase{\textit{et al.}}: A Sample Article Using IEEEtran.cls for IEEE Journals}

\IEEEpubid{}
% Remember, if you use this you must call \IEEEpubidadjcol in the second
% column for its text to clear the IEEEpubid mark.

\maketitle

% \begin{abstract}
% This document describes the most common article elements and how to use the IEEEtran class with \LaTeX \ to produce files that are suitable for submission to the IEEE.  IEEEtran can produce conference, journal, and technical note (correspondence) papers with a suitable choice of class options. 
% \end{abstract}
\input{sec/0_abstract}


% \section{Introduction}
% \IEEEPARstart{T}{his} file is intended to serve as a ``sample article file''
% for IEEE journal papers produced under \LaTeX\ using
% IEEEtran.cls version 1.8b and later. The most common elements are covered in the simplified and updated instructions in ``New\_IEEEtran\_how-to.pdf''. For less common elements you can refer back to the original ``IEEEtran\_HOWTO.pdf''. It is assumed that the reader has a basic working knowledge of \LaTeX. Those who are new to \LaTeX \ are encouraged to read Tobias Oetiker's ``The Not So Short Introduction to \LaTeX ,'' available at: \url{http://tug.ctan.org/info/lshort/english/lshort.pdf} which provides an overview of working with \LaTeX.
\input{sec/1_intro}

\input{sec/2_related_work}
\input{sec/3_method}

\input{sec/4_experiment}

\input{sec/5_conclusion}
\bibliographystyle{IEEEtran}
\bibliography{main}

\vfill

\end{document}

%% file: misc/__misc__.tex
\usepackage{amsmath,amsfonts}
\usepackage{algorithmic}
\usepackage{algorithm}
\usepackage{array}
\usepackage[caption=false,font=normalsize,labelfont=sf,textfont=sf]{subfig}
\usepackage{textcomp}
\usepackage{stfloats}
\usepackage{url}
\usepackage{verbatim}
\usepackage{graphicx}
\usepackage{cite}
\usepackage{booktabs}

\usepackage{xcolor, colortbl}
\usepackage{tabularx}
\usepackage{multirow}
\usepackage{enumitem}
\usepackage{bbm}
\usepackage{pifont}
\usepackage{capt-of}
\usepackage{array,multirow}
\definecolor{Gray}{gray}{0.85}
\definecolor{brown}{rgb}{0.8, 0.2, 0.2}
\definecolor{purple}{rgb}{0.65, 0.1, 0.75}
\definecolor{yellow}{rgb}{0.75, 0.75, 0.0}
\definecolor{orange}{rgb}{1.0, 0.5, 0.2}
\definecolor{green}{rgb}{0, 0.8, 0.2}
\definecolor{red}{rgb}{1.0, 0.0, 0.0}
\definecolor{dblue}{rgb}{0.6, 0.1, 0.9}

\definecolor{grey}{rgb}{0.9, 0.9, 0.9}
\newcommand{\ccol}{\cellcolor{grey}}

\newcommand{\bcmark}{\ding{51}}%
\usepackage[hidelinks,breaklinks]{hyperref}

%% file: sec/0_abstract.tex
\begin{abstract}
The core of cross-modal matching is to accurately measure the similarity between different modalities in a unified representation space. However, compared to textual descriptions of a certain perspective, the visual modality has more semantic variations. So, images are usually associated with multiple textual captions in databases. 
Although popular symmetric embedding methods have explored numerous modal interaction approaches, they often learn toward increasing the average expression probability of multiple semantic variations within image embeddings. Consequently, information entropy in embeddings is increased, resulting in redundancy and decreased accuracy.
In this work, we propose a Dynamic Visual Semantic Sub-Embeddings framework (\textbf{DVSE}) to reduce the information entropy. Specifically, we obtain a set of heterogeneous visual sub-embeddings through dynamic orthogonal constraint loss. To encourage the generated candidate embeddings to capture various semantic variations, we construct a mixed distribution and employ a variance-aware weighting loss to assign different weights to the optimization process. In addition, we develop a Fast Re-ranking strategy (\textbf{FR}) to efficiently evaluate the retrieval results and enhance the performance. We compare the performance with existing set-based method using four image feature encoders and two text feature encoders on three benchmark datasets: MSCOCO, Flickr30K and CUB Captions. We also show the role of different components by ablation studies and perform a sensitivity analysis of the hyperparameters. The qualitative analysis of visualized bidirectional retrieval and attention maps further demonstrates the ability of our method to encode semantic variations.

\end{abstract}

% 是不是叫子嵌入比较好 
% 语义变化 信息熵怎么联系起来 又可以称之为semantic ambiguity
% 出现的两种问题的命名 这个可以只用集合崩塌将二者关联起来，即集合崩塌发生在两种情况下：one-to-one, one-to-many，而第二种通常是欠缺考虑的。
% 解构这个词使用的是否恰当 不叫解构了，叫学习或者编码

%% file: sec/1_intro.tex
\section{Introduction}
\label{sec:intro}
% 香农借鉴了热力学的概念，将信息中排除冗余后的平均信息量称为“信息熵”[1]。在表示学习中，信息熵可以用来衡量潜在空间中嵌入变量的不确定性。由于自然界的图像通常包含大量像素，像素之间的排列方式可提供丰富的信息。这种复杂性和多样性导致视觉模态的信息熵相对较高，通常包含更多的语义变化。相反，文本由离散且有限的元素组成，通常更加结构化，在一般的检索数据集中具有明确的语义指向性，其语义变化较少。因此不同模态的潜在表示需要考虑不同程度的语义变化，其嵌入的信息熵也并不相等。
\IEEEPARstart{I}{mage}-text retrieval is the important component of multimodal learning and has been widely integrated into search engines in different domains\cite{gui2013performance}. 
\begin{figure}[ht]
\includegraphics[width=1.0\linewidth]{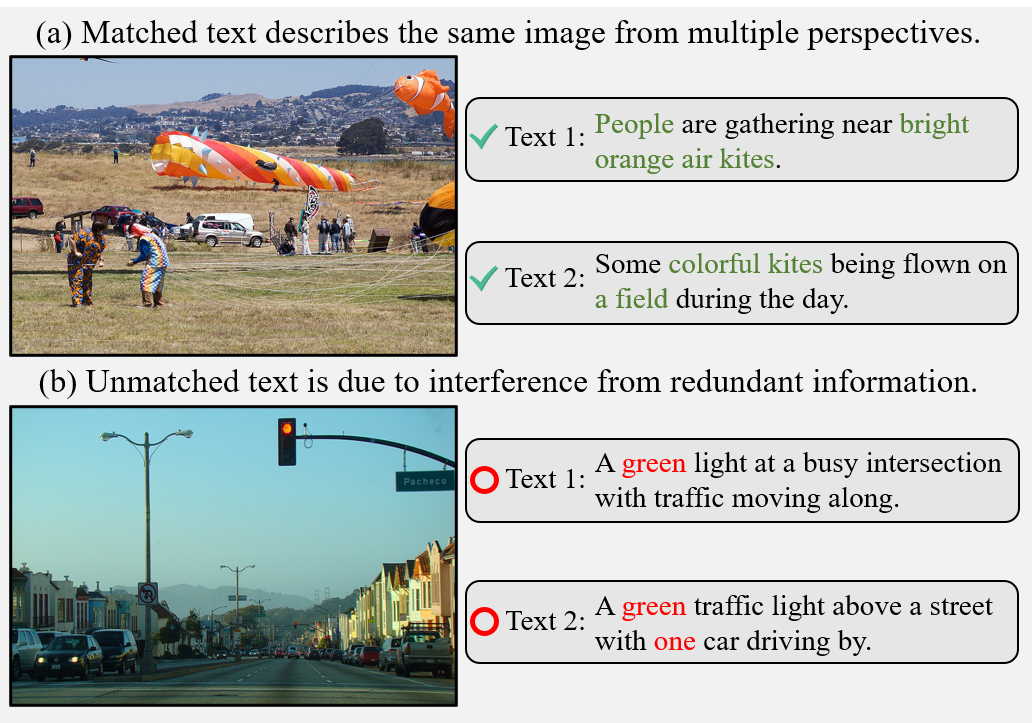}
\caption{Matched and unmatched image-text pairs on MSCOCO. (a) are pairs correctly matched and the words related to the key semantics are colored green. (b) are the mismatched pairs, where the words related to the wrong semantic are colored red.}
\label{fig: illustration}.
\end{figure}
It involves retrieving data from different modalities that have a relatively high semantic similarity to a given query image or text. However, diverse combinations of pixels in an image usually lead to more semantic variations \cite{li2022multi} than text. This poses a challenge to the retrieval process.

Approaches of image-text retrieval can be divided into two main categories based on the granularity of representation. The first type utilizes coarse-grained methods with a dual encoder  \cite{faghri2017vse++,frome2013devise,karpathy2015deep,qu2020context,karpathy2014deep,huang2018learning,kiros2014unifying,qu2021dynamic,chen2021learning}. By mapping different modal data to a unified metric space \cite{peng2022clustering, wang2019multi}, these methods can quickly calculate the similarity between the modalities. The second type utilizes fine-grained methods with interactive mechanisms \cite{lee2018stacked,chen2020imram,zhang2022negative}, such as self-attention, aiming to capture inter-modality correlations from more granular data segments. These two approaches often employ symmetric embedding representations, i.e., predicting a deterministic embedding for both the entire image (or each region) and the entire text (or each word). However, because images contain more semantic variations than text, the use of symmetric embeddings further increases the information entropy (i.e. uncertainty) of the image embeddings. 
\IEEEpubidadjcol
This implies that image embeddings must find an average representation of different semantic variations, resulting in more redundant information and fewer details. In the scenario depicted in Fig. \ref{fig: illustration}, the two text captions in (a) correspond to the same image, but convey drastically different semantics. Image embedding, on the other hand, requires encoding this information simultaneously, which can increase the ambiguity of image semantics \cite{song2019polysemous, kim2023improving} and inhibit retrieval performance. For example, in the scene depicted in (b), the complex background, i.e., traffic at the intersection, and buildings and lampposts along the street, causes the model to ignore basic discriminative information, e.g., the color of the traffic lights and the number of vehicles, leading to unmatched results.

\begin{figure}[t]
\includegraphics[width=1.0\linewidth]{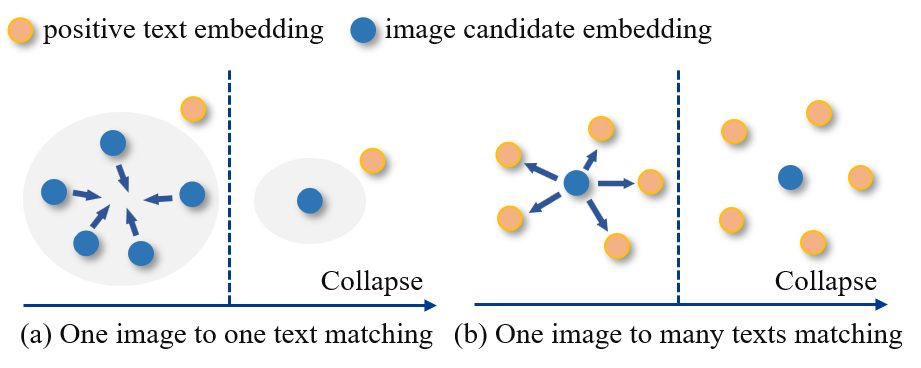}
\caption{illustration of the set collapse that occurs at the optimization process of one-to-one matching and one-to-many matching.}
\label{fig: set collapse}
\end{figure}

Recently, set-based models are proposed to encode semantic variations for more accurate matching. For example, CAMEAR \cite{qu2020context} and MVVSE \cite{li2022multi} aim to extract multiview features from images to explore within-class variations, while PVSE \cite{song2019polysemous} and DiVE \cite{kim2023improving} predict candidate embeddings for both images and text simultaneously to capture different semantics. However, these methods often face the challenges of set collapse. As shown in Fig. \ref{fig: set collapse}, it usually occurs in two cases: (a) in one-to-one matching, the optimized set lacks diversity and there is still redundant information between candidate embeddings; (b) in one-to-many matching, the candidate embedding is at the semantic center of multiple corresponding samples and fails to capture relevant semantic variations. Although the first case has received extensive attention, the second one has not been well studied, which may lead to the degradation of embeddings into high-entropy states.  

To address the aforementioned set collapse issues, this paper introduces a Dynamic Visual-Semantic Sub-Embeddings (\textbf{DVSE}) framework and a Fast Re-ranking (\textbf{FR}) strategy to achieve a generic representation of semantic variations for image-text retrieval. Specifically, we propose a dynamic orthogonal constraint loss to generate a set of heterogeneous image sub-embeddings. This loss ensures that the cardinality of the set can adaptively change while limiting the correlation between the sub-embeddings. It encourages the model to be able to take into account the effective training for all sub-embeddings and prevents one-to-one set collapse.
Meanwhile, we design a variance-aware weighting loss to address the issue of one-to-many set collapse. This loss leverages a mixed distribution constructed from non-target positive samples (i.e. positives that are not currently being optimized) and negative samples. The variance of this distribution is used as weights to motivate candidate embeddings to encode the most relevant semantics, thus suppressing embedding degradation. The devised Fast Re-ranking further enhances the model's resilience against noise. This strategy uses prior information from bidirectional retrieval to introduce a maximum discrimination term to the similarity matrix, for improving the retrieval performance.

% 具体而言，我们首先通过动态掩码和正交约束损失来生成a set of异构子嵌入。其中动态掩码保证集合基数能够自适应地改变，并鼓励模型关注不同子嵌入的优化过程。随后正交约束损失较为严格的限制了不同子嵌入在语义空间中的几何相关性，从而有效地防止intra-pair collapse。第二，我们介绍了一种方差约束损失，来解决inter-pair collapse的问题。该损失利用非目标正样本（目标正样本即当前优化的正样本）和负样本构建混合噪声分布，并平滑其方差作为不同候选嵌入的优化权重。不同的权重促使候选嵌入表达最相关的语义变化，从而阻止嵌入的退化。第三，为了进一步增强模型抵御噪声的能力，我们还设计了一种Fast Re-ranking strategy。该算法利用双向检索的先验信息，为相似度矩阵增加了最大值判别项，同时增强了检索的多样性和检索精度。
% First, to reduce the likelihood of non-target semantic variations (i.e., noise redundancy) in image embeddings, we employ a dynamic masking mechanism to predict a set of sub-embeddings. Second, an orthogonal constraint loss is applied to ensure that the geometric relationships of multiple sub-embeddings do not overlap in the semantic space. Third, we introduce a variance-aware constraint loss to assign different weights to the optimization process by constructing a mixed noise distribution. Different weights encourage different candidate embeddings to express the most relevant semantics, effectively preventing degradation. Finally, we propose a consistency assumption and design a Fast Re-ranking strategy to further reduce the noise of image-text embeddings.
In summary, the contributions of this paper can be summarized as follows.
\begin{itemize}
    \item We propose a Dynamic Visual Semantic Sub-Embeddings framework, which addresses the set collapse problem faced by set-based models and motivates the model to learn different semantic variations.
    \item We design a Fast Re-ranking strategy, which discards the nearest-neighbor search process used in previous methods, providing an efficient and robust refinement of the image-text similarity.
    \item We perform extensive experiments on datasets with heterogeneous data distributions, such as MSCOCO \cite{lin2014microsoft} and Flickr30k \cite{young2014image}, as well as the CUB Captions dataset with homogeneous data distribution \cite{wah2011caltech}. The results confirm the state-of-the-art performance of our method.
\end{itemize}
% Therefore, latent representations of different modalities should consider varying degrees of semantic variations, and the information entropy in their embeddings is not equal.

% 图像通常包含更多的语义变化，而文本仅从某一个视角进行描述。采用对称嵌入方法不仅降低了图像和某段匹配文本之间的相似性，同时也引入了噪声冗余。

%% file: sec/2_related_work.tex
\section{Related Work}

% \subsection{对称嵌入检索}
% 对称嵌入的方法根据交互模式的不同可以分为使用双编码器的全局粗粒度方法和使用自注意力等交互机制的局部细粒度方法。其中，双编码器的结构简单，分别利用独立的图像和文本编码器来提取数据的整体嵌入。而不同损失函数的设计将会确保匹配的查询-值对具有比非匹配对更高的相似度。

% 相关研究积极探索了模型结构[1][2][3]和损失函数[4][5][6]的改进。例如，Frome等人[4]使用了CNN作为视觉模型，Skip-Gram作为语言模型，并补充了基于三元组的hinge损失进行模型优化。Faghri等人[5]引入了在线硬负样本挖掘策略以进一步提高模型性能。Chen等人[6]提出了一个通用池化操作器，用于学习不同特征的最佳池化策略。然而，这一类结构因为缺少图像和文本交互的过程，通常很难捕获更细粒度的信息。近年来，基于局部信息的方法受到了广泛关注，例如使用交叉注意力结构来捕获图像显著区域与文本单词之间的对应关系[7][8][9]，或者专注于图神经网络的创建，以聚合邻域信息的对象关系元组[10][11][12][13]。随着大型模型的崛起，研究人员开始探索基于单个或双变换器的视觉-语言预训练(VLP)模型[14][15][16][17][18]，并设计了各种预训练任务，以增强模型的跨模态交互能力并提高模型的准确性。然而，这种方法通常需要消耗巨大的计算量。同时，我们注意到一些Re-ranking算法[1][2]也在不断兴起，他们通常以最近邻搜索为起点进一步挖掘图文双向检索的内在联系，一般具有较高的计算复杂度。

% \subsection{非对称嵌入检索}
% 尽管对称嵌入方法在模态表征和交互方式方面得到了深入研究，但它们忽视了不同模态天生包含不同语义变化的事实。单一的对称嵌入通常导致包含更多语义变化的模态使用信息熵更大的嵌入。为了解决这一问题，近年来，基于集合的嵌入方法受到了广泛关注。研究者试图将一个样本编码成一组异构的候选嵌入，以解构不同的语义变化。PVSE通过多头自注意力机制和残差学习，将全局和局部特征相结合，以计算实例的多个不同表示。PCME在此基础上将样本建模为嵌入空间中的概率分布，从而表示联合嵌入空间中的一对多关系。DiVE则利用插槽注意力机制改进了集合预测模型，并设计smooth-Chamfer similarity尝试解决稀疏监督和集合崩溃的问题。而CMAEAR和MVVSE则有所不同，它们更侧重于提取图像的候选嵌入集合，仅提取单个文本嵌入，以平衡不同模态之间信息密度的差异。

% 非对称嵌入方法扩展了嵌入向量中语义变化的编码空间，使模型能够专注于捕获目标语义变化，而不必提高多种语义变化的平均表达概率。然而，这些方法通常固定了嵌入集合的基数，无法自适应地调整，并且不同候选嵌入与待匹配样本间的优化权重是等价的，这可能会导致每个候选嵌入传达更加模糊的信息，使得它们退化为具有较高信息熵的状态。

% 考虑到信息密度的不同，本文提出的方法类似于[1][2]，仅提取图像模态的子嵌入集合，并通过动态掩码和正交约束损失来获取异构候选嵌入，具有更高的自适应性。同时，引入的方差约束损失可以根据信息熵的大小调整不同候选嵌入的优化方向，从而有效防止嵌入的退化。而Fast Re-ranking则会根据嵌入的信息熵重新评估计算结果，且计算简单高效。

\subsection{Symmetric Embedding Retrieval}
Methods based on symmetric embeddings can be categorized into two types upon the granularity of representation: coarse-grained global methods and fine-grained local methods. The global methods \cite{faghri2017vse++,frome2013devise,karpathy2015deep,qu2020context,karpathy2014deep,huang2018learning,kiros2014unifying,qu2021dynamic,chen2021learning} have a straightforward structure, employing separate encoders for images and texts to extract global features from the data. Besides, the design of loss functions \cite{chen2020adaptive,frome2013devise,wei2021universal,faghri2017vse++} ensures that the matched query-value pairs exhibit higher similarity than unmatched pairs. For example, Frome et al. \cite{frome2013devise} use CNN as the visual model, Skip-Gram as the language model, and supplemented with hing-based triplet loss for model optimization. Faghri et al. \cite{faghri2017vse++} introduce an online hard negative mining strategy to further improve the model performance. Chen et al. \cite{chen2021learning} propose a Generalized Pooling Operator to learn the best pooling strategy for different features. Local methods have also gained attention due to their strong interaction capabilities, such as cross-attention structures that are used to capture correspondences between salient regions in images and text words \cite{lee2018stacked,chen2020imram,zhang2022negative}, and the creation of graph neural networks aggregate object relation tuples in neighborhood information \cite{liu2020graph,li2019visual,wang2020cross,diao2021similarity}. Lee et al. first \cite{lee2018stacked} introduce a stacked cross-modal attention mechanism to discover complete latent visual-semantic alignments. Chen et al. \cite{chen2020imram} employ an iterative matching scheme to progressively explore fine-grained cross-modal correspondences. Li et al. \cite{li2019visual} use graph convolutional networks to capture key objects and semantic concepts in the scene. 
With the rise of large-scale models, researchers have begun exploring transformer-based visual language pre-training (VLP) models \cite{lu2019vilbert,chen2020uniter,li2021align,kim2021vilt,ji2023map} and have designed various pre-training tasks to improve the retrieval precision of the models, including Masked Language Modeling (MLM), Image-Text Matching (ITM), and Vision-Language Contrastive Learning (VLC), etc.

Unfortunately, due to the limitations of dual-encoder structure, global methods encode semantic variations within a single embedding. This often increases retrieval uncertainty and reduces accuracy. Although local methods and VLP methods attempt to link the corresponding relations between salient regions and key words, further reducing the information entropy of embedding poses higher demands on the granularity of information and computational resources.
% In reality, semantic variations exist in the multiple cognitions of macro scenes. SGRAF \cite{diao2021similarity}, on the other hand, infers relationships and similarities in local and global alignment through similarity graph reasoning and attention filtering networks.
\begin{figure*}[t]
\includegraphics[width=1.0\linewidth]{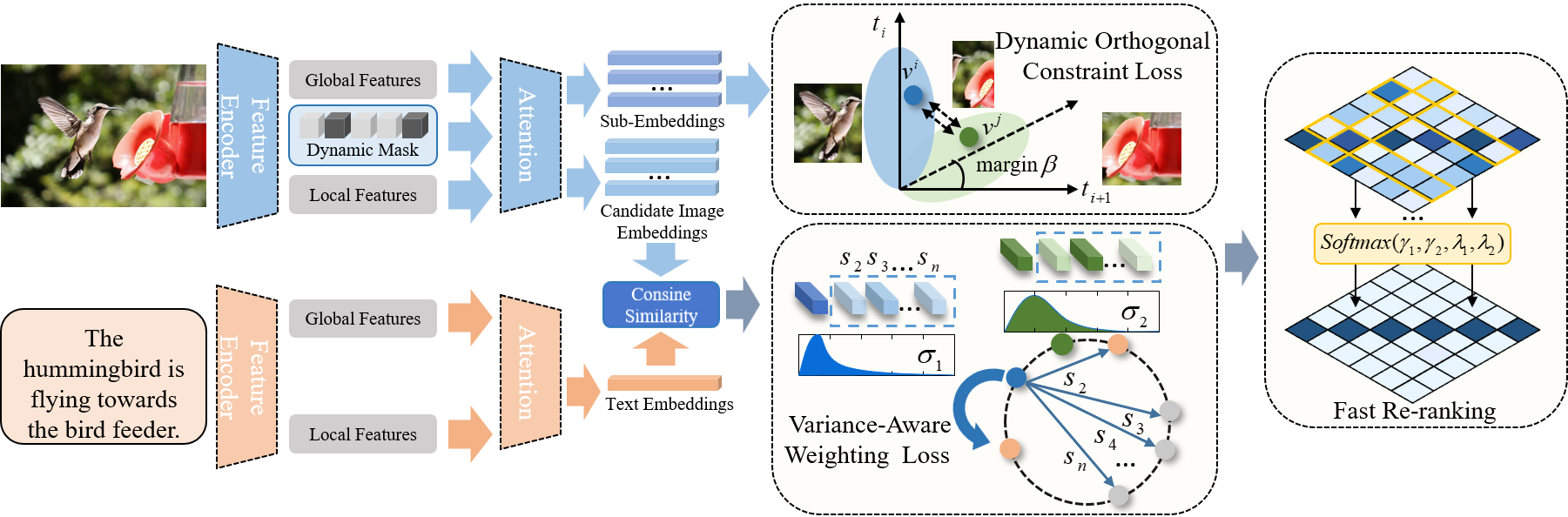}
\caption{The overview of our model. Raw data are initially processed through a feature encoder to obtain local and global features. Subsequently, both features are fused into the predicted embedding using a self-attention module. Simultaneously, local features in image modality are also used to generate a set of sub-embeddings via multi-head self-attention and dynamic mask, while the dynamic orthogonal constraint loss is computed. Taking into account the differences in information density, we extract a set of candidate embeddings from images, while only a single predicted embedding is extracted from text. Finally, we calculate the variance-aware weighting loss on the similarity matrix of image and text, and employ FR during testing.}
\label{fig: main}
\end{figure*}
% We use a dual-encoder structure, where the raw data is first processed by a feature extractor, resulting in local features that are then fed into two distinct branches. The first branch produces global features through the max-pooling operation. The second branch processes local features through a multi-head self-attention and then connects them to the global features via residual connections, producing the final predicted embeddings. A dynamic mask is applied to the outputs of the multi-head attention to select effective sub-embeddings. Taking into account the variance in information density, we extract a set of sub-embeddings from the images and only one sub-embedding from the text.
\subsection{Asymmetric Embedding Retrieval}
Symmetric methods have explored modal representation and interaction, but have not taken into account that different modalities contain varying semantic variations. This often leads to higher information entropy in modality embeddings. In recent years, set-based methods \cite{song2019polysemous,chun2021probabilistic,kim2023improving,qu2020context,li2022multi,wang2020cross} have tried to solve this problem. They encode a sample into a set of candidate embeddings to capture different semantic variations and focus on how to effectively train elements within a set. Song et al. \cite{song2019polysemous} combine global and local features using multi-head self-attention and residual learning to compute multiple representations of an instance. Kim et al. \cite{kim2023improving} improve the set prediction model using slot attention and introduce smooth-Chamfer similarity to address sparse supervision and set collapse issues. On the other hand, to balance the differences in information density between different modalities, Qu et al. \cite{qu2020context} and Li et al. \cite{li2022multi} extract candidate embedding sets from images, while only extracting a single text embedding. These methods have successfully alleviated the issue of set collapse in one-to-one matching through dense supervision. However, little research has specifically addressed set collapse in one-to-many matching, and the equal optimization weights for different candidate embeddings increase their semantic ambiguity. 

Methods based on probability or geometric embeddings are proposed to model the ambiguities inherent in the modality. For example, Chun et al. \cite{chun2021probabilistic} construct a richer embedding space to implicitly capture the one-to-many associations. Wang et al. \cite{wang2022point} introduce a geometric representation of points-to-rectangles and query more relevant points within the rectangles. Li et al. \cite{li2022differentiable} directly optimize diversity metrics through differentiable approximation functions. Upadhyay et al. \cite{upadhyay2023probvlm} estimate the probability distribution of pre-trained VLM embeddings in a posterior manner, avoiding the need to retrain large models from scratch. These methods perform uncertainty estimation and expand the potential set of results. However, they do not explicitly model and reduce the embedding entropy, and redundant information still exists in the modality caused by semantic variations.

\subsection{Re-ranking strategy}
Re-ranking, which is widely applied in domains such as re-identification \cite{garcia2015person,zhong2017re,liao2023multi} and image retrieval \cite{tan2021instance,zhang2020understanding,radenovic2018fine}, utilize high-confidence samples to reorder the initial retrieval results. Based on the interactive nature of bidirectional retrieval in cross-modal matching, re-ranking strategies refine the retrieval results by performing reverse retrieval on elements in the similarity matrix \cite{wei2020boosting,wang2019matching,yuan2022remote,wang2023dual}. Among them, Wang et al. \cite{wang2019matching} introduce the fundamental assumption of reverse retrieval and narrow the gap between training and testing by searching for the nearest neighbors of k reciprocally. Yuan et al. \cite{yuan2022remote} further explore the similarity matrix, and optimize the retrieval results using multiple ranking factors. Wang et al. \cite{wang2023dual} investigate reciprocal contextual information and refine coarse retrieval results through neighbor-related sorting. However, these methods rely on nearest neighbor search to further explore the underlying connections in bidirectional image-text retrieval, often accompanied by relatively high runtime.

In this paper, to reduce information entropy in embedding and eliminate ambiguity, we encode semantic variations with dynamic sub-embeddings based on the set-based model. Considering the difference in information density, we extract multiple sub-embeddings for images and use different weights to guide heterogeneous embeddings to capture semantics. Such a dynamic sub-embeddings method not only prevents one-to-one set collapse, but also solves the one-to-many set collapse problem. Meanwhile, in order to speed up the re-ranking and enhance the noise resistance of the model, we design a Fast Re-ranking strategy by performing bidirectional normalization on the similarity matrix. Compared with nearest neighbor search, this strategy is efficient while having higher accuracy.

%% file: sec/3_method.tex
\section{Methodology}
The architecture of our method is illustrated in Fig. \ref{fig: main}. In Section 3.1, we discuss the background. Section 3.2 demonstrates how to build dynamic orthogonal constraint loss and generate the final predicted embeddings.  The Variance-aware weighting Loss and the Fast Re-Ranking strategy are described in Sections 3.3 and 3.4, respectively.

\subsection{Background}
For a visual language training set $\mathcal{D} = (\mathcal{X},\mathcal{Y})$, traditional symmetric embedding techniques employ two embedding functions, $\Phi_{\mathcal{V}}$ and $\Psi_{\mathcal{T}}$, to map image and text samples into the embedding space $\mathbb{R}$. The similarity between two samples is then determined by the distance between the two vectors in this space. In this process, convolutional neural networks or pre-trained object detectors are commonly used as image feature extractors $g_{\mathcal{V}}(\cdot)$. For an image $x \in \mathcal{X}$, the set of local visual features is defined as $\mathbf{z}_{v} = g_{\mathcal{V}}(x) \in \mathbb{R}^{R\times D}$, $R$ is the number of regions, and after a pooling operation $h_{\mathcal{V}}(\cdot)$, the global feature is aggregated as $\mathbf{v} = h_{\mathcal{V}}(\mathbf{z}_v) \in \mathbb{R}^{D}$. Similarly, for text data, sequence models $g_{\mathcal{T}}(\cdot)$ are used to extract token features. For a caption $y \in \mathcal{Y}$, the sequence of token features is denoted as $\mathbf{z}_{t} = g_{\mathcal{T}}(y) \in \mathbb{R}^{L\times D}$, $L$ is the length of the caption, and the global feature $\mathbf{t} = h_{\mathcal{T}}(\mathbf{z}_t) \in \mathbb{R}^{D}$ is then generated after a pooling operation $h_{\mathcal{T}}(\cdot)$ .

We formulate the matching objective between images and texts using the softmax criterion. Assuming that there are $n$ captions and $n$ images in the dataset, with the set $\mathbb{C}$ containing $m$ captions that match the image $x$, the probability of the correct match between a caption $y_{i} \in \mathbb{C}$ and the image $x$ can be expressed as:
\begin{equation}
P(\Theta|\mathbf{v},\mathbf{t}_{i}) = \frac{exp((\mathbf{t}_{i}^{T}\mathbf{v})/\sigma^{2})}{\sum_{j=1}^{n}exp((\mathbf{t}_{j}^{T}\mathbf{v})/\sigma^{2})}
\label{eq: 1}
\end{equation}
The temperature $\sigma$ is considered, and both $\mathbf{v}$ and $\mathbf{t}_j$ are limited to the $\ell_2$-normalized unit sphere. By default, $\Theta$ is set to 1, which implies that the sample belongs to $\mathbb{C}$. If its value is 0, then the sample is from the noise distribution. For an image embedding, which typically contains semantic variations corresponding to multiple textual descriptions as well as the irrelevant noise information. Taking $\mathbf{t}_{i}$ as different semantic variations to the image; we can calculate the information entropy of the image embedding:
\begin{equation}
H(\mathbf{v}) = -\sum_{i=1}^{m}P(\Theta|\mathbf{v},\mathbf{t}_{i})\log P(\Theta|\mathbf{v},\mathbf{t}_{i})-P(\xi)\log P(\xi)
\end{equation}
the noise is represented by $\xi$. To match texts of different semantics, the model will increase the average probability of multiple semantic variations, causing $H$ to increase with $m$. The purpose of this paper is to use image sub-embeddings to decompose this process, allowing each sub-embedding to focus on capturing specific semantic variations.
\subsection{Dynamic Orthogonal Sub-Embeddings}
In this section, we introduce the design of sub-embeddings and suggest a dynamic orthogonal constraint loss to reduce the correlation between them, thus emphasizing the desired semantic variations. Considering the difference in information density, we extracted sub-embeddings only for images.
\begin{figure}[tbp]
\includegraphics[width=1.0\linewidth]{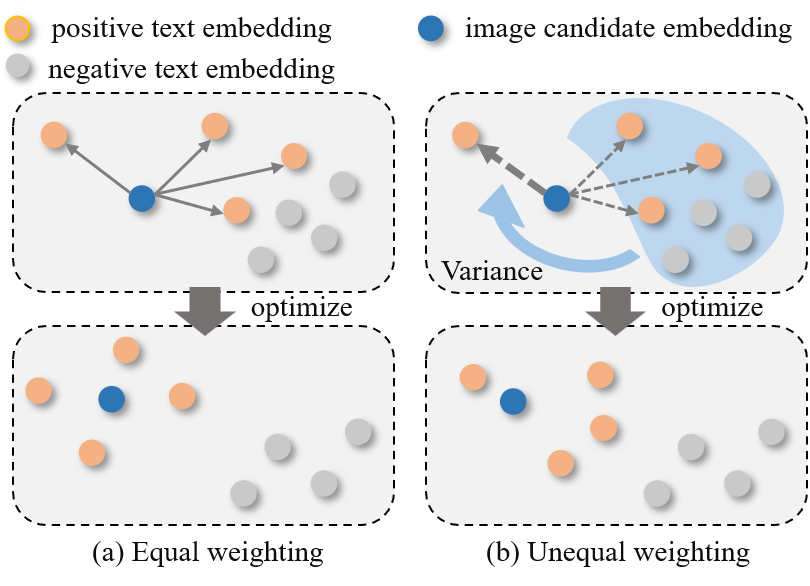}
\caption{Comparison between the equal weighting and unequal weighting. The solid line indicates the model with the same weights, while the dashed line shows different weights. The blue area represents the mixed distribution (including non-target positive samples and negative samples). We use variance-aware weighting loss to calculate the variance of the mixed distribution as weight, thus motivating the image candidate embedding to concentrate on capturing the relevant semantic variation.}
\label{fig: Lvar}
\end{figure}

\noindent
\textbf{Embeddings Prediction:} We adopt multi-head self-attention \cite{vaswani2017attention} for sub-embedding extraction \cite{song2019polysemous, chun2021probabilistic} for visual modality. Specifically, for an image $x$, its local information is encoded into a sub-embedding set containing $K$ elements. Each element is obtained through $\hat{\mathbf{v}}^{k} = s(w^{1}att_{\mathcal{V}}^{k}(\mathbf{z}_{v})\mathbf{z}_{v}), k \in \{1,\cdots,K\}$, where $s$ is an activation function, $w^{1}$ are parameters of a fully connected layer, and $att_{\mathcal{V}}^{k}$ denotes different attention heads. The final candidate visual embedding is then obtained by residual connection of global and local information: $\mathbf{v}^{k} = LN(h_{\mathcal{V}}(\mathbf{z}_{v})+\hat{\mathbf{v}}^{k})$, where $LN$ is the LayerNorm \cite{ba2016layer}. For the text modality, we do not predict sub-embeddings but instead directly use self-attention to output the final embedding: $\mathbf{t} = LN(h_{\mathcal{T}}(\mathbf{z}_{t})+s(w^{3}att_{\mathcal{T}}(\mathbf{z}_{t})\mathbf{z}_{t}))$.

\noindent
\textbf{Dynamic Orthogonal Constraint Loss:} 
We need to ensure that sub-embeddings associate different semantic variations to prevent one-to-one set collapse. However, a set-based model often falls into a local minimum due to the large number of sub-embeddings that are not optimized \cite{li2022multi}. Thus, simply penalizing the correlation between sub-embeddings is not enough to help them learn meaningful semantic variations. In this paper, we add a dynamic masking mechanism to adaptively select image sub-embeddings by aggregating local visual features.
\begin{equation}
\mathcal{F}_{\text{mask}} = round(s(h_{\mathcal{V}}(w^{2}\mathbf{z}_v)))
\end{equation}
where $round(\cdot)$ represents rounding operation. We keep the fully connected layer with the weight $w^{2}$ fixed and only use it for dimension reduction. At this stage, we generate different masks for different images. The model cannot rely on a particular sub-embedding to learn all the information. So, we keep the orthogonality between the different sub-embeddings by a dynamic orthogonal constraint loss:
\begin{equation}
\mathcal{L}_{dc} = \left[\sum_{i=1}^{K}\sum_{j=1,j \neq i}^{K}||\mathcal{F}_{mask}((\hat{\mathbf{v}}^{i})^{T})\mathcal{F}_{mask}(\hat{\mathbf{v}}^{j})||-\beta\right]_{+}
\label{eq: loss_ort}
\end{equation}
The $\beta$ is used to modify the boundary of the correlation between sub-embeddings. Too strict orthogonality may lead to inadequate embedding representations, we incorporate this parameter to preserve partial of the semantic relations. The hinge function $[\cdot]_{+}$ denotes $max(x, 0)$.

% （放在背景里或者本段前面介绍）Constraining the embedding vectors to be orthogonal to each other is a commonly used de-correlation method \cite{hazarika2020misa, song2019polysemous}, and together with dynamic masking, the model is able to address the one-to-one set collapse while optimising all the sub-embeddings at the same time.

\begin{figure}[t]
\includegraphics[width=1.0\linewidth]{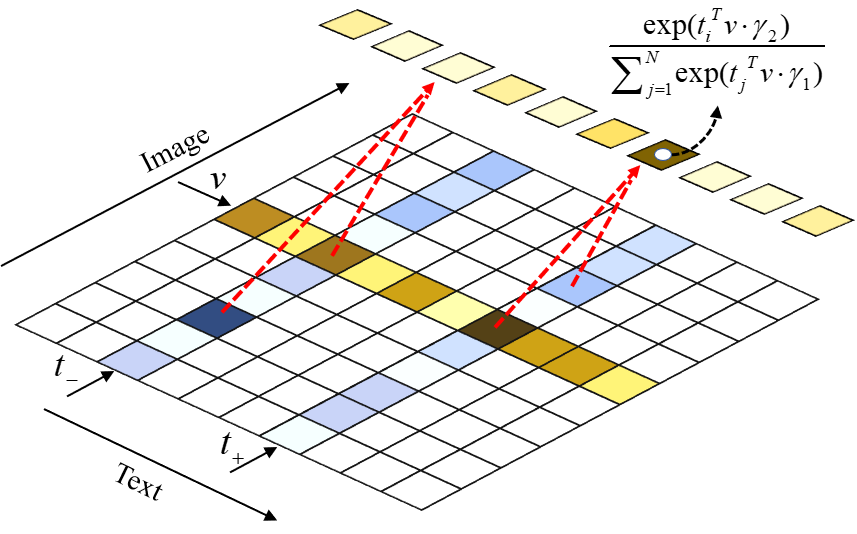}
\caption{
Illustration of Fast Re-rank in row-wise retrieval: when compare the similarity of two elements in a row, we normalize the column in which the element is located, thus introducing the information from the column-wise retrieval into the sorting process of the row-wise retrieval.}
\end{figure}

\subsection{Variance-Aware Weighting Loss}
\noindent
\textbf{Variance weighting:} Previous methods did not consider the set collapse in one-to-many matching. As shown in Fig. \ref{fig: Lvar}(a), they typically used the same weights in the optimization process, which could lead to degradation of the embeddings to a state of high information entropy. Conversely, if we assign different weights to candidate embeddings, it will facilitate them to encode different semantic variations Fig. \ref{fig: Lvar}(b). So, we reformulate Equation \ref{eq: 1} by replacing the original image embedding with candidate embeddings:
\begin{equation}
P(\Theta|\mathbf{v}^{k},\mathbf{t}_{i}) = \frac{exp((\mathbf{t}_{i}^{T}\mathbf{v}^{k})/\sigma^{2}_{k})}{\sum_{j=1}^{n}exp((\mathbf{t}_{j}^{T}\mathbf{v}^{k})/\sigma^{2}_{k})}
\end{equation}
Considering that $\sigma^{2}$ in the equation is for regulating the smoothness of the similarity distribution, when we adopt the variance of the similarity distribution as $\sigma^{2}$, the smaller variance reflects higher confidence, which in turn encourages different candidate embeddings to express the most relevant semantics. 
\begin{equation}
    \sigma_{k}^{2} = (Std(P(\Theta|\mathbf{v}^{k},\mathbf{t}_{i}))+1)^{2}
\end{equation}
where $Std(\cdot)$ represents the sample standard deviation of the similarity distribution, and to prevent $\sigma_{k}$ from becoming too small, we apply an additive one operation to it. Since set-based models predict only deterministic values for each candidate embedding and do not possess the distribution information, we treat the remaining samples in a batch as results sampled from a mixed distribution $\hat{P}(\Theta=0)$:
\begin{equation}
    \hat{P}(\Theta=0)=P(\Theta=0)+P(\Theta=1|\mathbf{v}^{k},\mathbf{t}_{j,j \neq i})
\end{equation}
The right-hand of the equation consists of two terms: the first is the distribution of negative samples, while the second is the distribution of the positive sample set $\mathbb{C}$ with the $i$-th element removed. This mixed distribution treats all results outside the semantic variation corresponding to the candidate embedding as negative samples. The variance of this distribution reflects the model's ability to recognize none-target semantics, with higher variance indicating lower confidence of the candidate embedding. Then the weights can be calculated as follows:
\begin{equation}
    \sigma_{k}^{2} = (Std(\hat{P}(\Theta=0))+1)^{2}
\end{equation}
% Illustration of Fast Re-rank: Through the normalization of each row and each column of the similarity matrix, the model reassess the results of row retrieval (or column retrieval) by considering the maximum values from column retrieval (or row retrieval)

\noindent
\textbf{Objective Function:} After obtaining the weights, we incorporate all candidate embeddings into the calculation of our objective function:
\begin{equation}
\begin{aligned}
    \mathcal{L}_{var}(x,y_{i}) &= -log(\prod_{k=1}^{K}P(\Theta|\mathbf{v}^{k},\mathbf{t}_{i})) \\
    &= -log \prod_{k=1}^{K}\frac{exp((\mathbf{t}_{i}^{T}\mathbf{v}^{k})/\sigma^{2}_{k})}{\sum_{j=1}^{n}exp((\mathbf{t}_{j}^{T}\mathbf{v}^{k})/\sigma^{2}_{k})} \\
    &=\sum_{k=1}^{K}(\frac{1}{\sigma_{k}^{2}}\mathcal{L}^{'}_{k}+log\frac{\sum_{j=1}^{n}exp((\mathbf{t}_{j}^{T}\mathbf{v}^{k})/\sigma^{2}_{k})}{(\sum_{j=1}^{n}exp(\mathbf{t}_{j}^{T}\mathbf{v}^{k}))^{\frac{1}{\sigma^{2}_{k}}}}) \\ &\approx \sum_{k=1}^{K}(\frac{1}{\sigma_{k}^{2}}\mathcal{L}^{'}_{k}+log\sigma_{k})
\end{aligned}
\end{equation}
where $\mathcal{L}^{'}_{k}$ is defined as $-\log\frac{exp(\mathbf{t}_{i}^{T}\mathbf{v}^{k})}{\sum_{j=1}^{n}exp(\mathbf{t}_{j}^{T}\mathbf{v}^{k})}$. In the final transition term, we make an assumption similar to \cite{kendall2018multi} to simplify $\frac{1}{\sigma_{k}}\sum_{j=1}^{n}exp((\mathbf{t}_{j}^{T}\mathbf{v}^{k})/\sigma^{2}_{k}) \approx (\sum_{j=1}^{n}exp(\mathbf{t}_{j}^{T}\mathbf{v}^{k}))^{\frac{1}{\sigma^{2}_{k}}}$, when $\sigma_{k} \rightarrow 1$, the equality is satisfied. In practice, for $\mathcal{L}^{'}_{k}$, we employ the widely used hard sample mining strategies \cite{faghri2017vse++}:
\begin{equation}
\begin{aligned}
    \mathcal{L}^{'}_{k} &= log(\mathop{max}_{j}[exp((\mathbf{t}_{j}^{T}\mathbf{v}^{k}) - (\mathbf{t}_{i}^{T}\mathbf{v}^{k}) + \alpha)]_{+}) \\
    &= \mathop{max}_{j}[\mathbf{S}(\mathbf{v}^{k},\mathbf{t}_j)-\mathbf{S}(\mathbf{v}^{k},\mathbf{t}_i) + \alpha]_{+}
\end{aligned}
\end{equation}
where $\mathbf{S}(\cdot)$ is the cosine similarity. For bidirectional image-text retrieval, our variance-aware weighting loss function is defined as:
\begin{equation}
\begin{aligned}
    \mathcal{L}_{var} &= \sum_{k=1}^{K}\sum_{i=1}^{n}(\frac{1}{\sigma^{2}_{k}}\mathop{max}_{j}[\mathbf{S}(\mathbf{v}^{k}_{i},\mathbf{t}_j)-\mathbf{S}(\mathbf{v}^{k}_{i},\mathbf{t}_i) + \alpha]_{+} + log\sigma_{k}) \\ &+ \sum_{k=1}^{K}\sum_{i=1}^{n}(\frac{1}{\sigma^{2}_{k}}\mathop{max}_{j}[\mathbf{S}(\mathbf{t}_{i},\mathbf{v}^{k}_j)-\mathbf{S}(\mathbf{t}_{i},\mathbf{v}^{k}_i) + \alpha]_{+} + log\sigma_{k})
\end{aligned}
\label{eq: loss_var}
\end{equation}
Finally, we introduce a scaling factor $\eta$ to balance the dynamic orthogonal constraint loss and variance-aware weighting loss:
\begin{equation}
    \mathcal{L} = \eta \mathcal{L}_{var} + (1 - \eta) \mathcal{L}_{ort}
\label{eq: loss_final}
\end{equation}

\input{table/comparison_result_about_MSCOCO}

\subsection{Fast Re-ranking}
% 通过进一步挖掘相似度矩阵的信息来对原始结果重新评估,re-rank算法在过去的图文检索当中取得了富有成效的结果[1][2]。然而他们通常将这一过程看作为最近邻搜索问题，需要复杂的计算过程。本文的方法则极大地简化了这些步骤，直接作用于相似度矩阵。具体而言：我们利用了图像文本嵌入的信息熵，提出一个一致性假设即：在测试阶段。对于行列检索（即文本和图像检索）输出的相似度矩阵，其中每一个元素的行列检索结果的置信度应当是一致的。基于这一假设，我们提出了跨模态一致性约束策略。我们以行检索为例，对于模型输出的相似度矩阵$\boldsymbol{A}$，我们对其中的任一元素$\boldsymbol{A}_{i.j}$进行列归一化:
% To further reduce the information entropy of image and text embeddings, we propose \textbf{a} \textbf{consistency} \textbf{assumption}: during testing, for the similarity matrix generated in both row-wise and column-wise retrievals (i.e., text and image retrieval), the confidences of the results for each element in the matrix should be consistent. This means that embeddings with higher similarity to negative samples in the column-wise retrieval exhibit higher uncertainty and need re-evaluation in the row-wise retrieval, and vice versa. Based on this assumption, we introduce a cross-modal consistency constraint strategy. Taking row-wise retrieval as an example, for any element $A_{i,j}$ in the similarity matrix $\boldsymbol{A}$ generated by the model, we normalize the columns to obtain the following:
 To further improve the model performance, re-ranking strategies have been adopted in image-text retrieval tasks by exploring the constraint relations between bidirectional retrieval. However, these methods often treat this process as a nearest neighbor search problem that involves complex computations. To integrate the bidirectional retrieval information more efficiently, we normalize the similarity matrix instead. Based on the consistency assumption \cite{wang2019matching}, the confidence of an element in both row-wise and column-wise retrievals (i.e., text and image retrieval) should be consistent during testing. Therefore, both text and image retrieval can be mutually constrained. For the generated similarity matrix $\boldsymbol{A}$, we perform column and row normalization on all elements $\boldsymbol{A}_{i,j}$. Taking row-wise retrieval as an example, we use LogSumExp (LSE) to approximate the maximum function to introduce more discriminating information:
\begin{equation}
    \begin{aligned}
        \tilde{A}_{i,j} &= \frac{exp(A_{i,j})}{\sum_{l}exp(A_{l,j})} \\
        &= (1+exp(log(\sum_{l,l \neq j}exp(A_{l,j}))-A_{i,j}))^{-1} \\
        &= (1+exp(LSE\{A_{l,j} \in \boldsymbol{A}_{i,:}|l \neq i\}-A_{i,j}))^{-1} \\
        &\leq (1+exp(\mathop{max}\{A_{l,j} \in \boldsymbol{A}_{i,:}|l \neq i\}-A_{i,j}))^{-1} 
    \end{aligned}
\label{eq: normalize}
\end{equation}
where $\tilde{A}_{i,j}$ is the normalized element. In this process, each element is compared with the maximum value in its own column. Every column reflects the similarities between a text embedding and all image embeddings. If $A_{i,j}$ is a positive sample, the lower the maximum value (negative sample) in its column, the higher the confidence of the text embedding, and then $\tilde{A}_{i,j}$ has a relatively higher value. In contrast, if $A_{i,j}$ is a negative sample, the higher the maximum value (postive sample) in its column, the higher the confidence in text embedding, resulting in a relatively lower value for $\tilde{A}_{i,j}$. This re-ranking method allows us to effectively use the prior information from the original matrix (i.e. the maximum values in columns) without bothering a nearest-neighbor search process. Additionally, we add scale parameters to the normalization process to make the boundary of the Equation \eqref{eq: normalize} tighter.
\begin{equation}
    \begin{aligned}
        \tilde{A}_{i,j} &= (\frac{\sum_{l}exp(A_{l,j}\cdot\gamma_{1})}{exp(A_{i,j}\cdot\gamma_{2})})^{-1} \\
        &< (1+exp(\gamma_{1}\cdot\mathop{max}\{A_{l,j} \in \boldsymbol{A}_{i,:}|l \neq i\}-\gamma_{2}\cdot A_{i,j}))^{-1} 
    \end{aligned}
\end{equation}
where the parameter $\gamma_{1}$ is used to adjust the tightness of the inequality, while $\gamma_{2}$ emphasizes the contribution of $A_{i,j}$ in the normalization. 
Finally, we output two similarity matrices for text and image retrieval, respectively:
\begin{equation}
\begin{aligned}
    &\mathbf{A}_{i-t} = softmax0(\mathbf{A}, \gamma_{1}, \gamma_{2}) \\
    &\mathbf{A}_{t-i} = softmax1(\mathbf{A}, \lambda_{1}, \lambda_{2})
\end{aligned}
\label{eq: normal param}
\end{equation}
\input{table/comparison_result_about_flickr30k}
where $softmax0(\cdot)$ and $softmax1(\cdot)$ refer to the normalization of the matrix along columns and rows.

%% file: table/comparison_result_about_MSCOCO.tex
\begin{table*}[t]
\centering
\caption{
The performance comparison between our model and other approaches on the MSCOCO dataset. We present the Recall@K and RSUM metric results, including both the 1K test setting (averaged over 5-fold test datasets) and the 5K test setting. The best results are highlighted in bold.
}
% \scalebox{1}{
\footnotesize
\setlength{\tabcolsep}{5pt}
\begin{tabular}{l|ccc|ccc|c|ccc|ccc|c}
\toprule
    \multicolumn{1}{l}{\multirow{3}{*}[-1.6em]{Method}} 
    & \multicolumn{7}{c|}{1K Test Images} 
    & \multicolumn{7}{c}{5K Test Images} \\ \midrule
    \multicolumn{1}{c|}{}
    % & 
    & \multicolumn{3}{c|}{Image-to-Text} 
    & \multicolumn{3}{c|}{Text-to-Image}
    & \multicolumn{1}{c|}{\multirow{2}{*}{\textsc{RSUM}}}
    & \multicolumn{3}{c|}{Image-to-Text} 
    & \multicolumn{3}{c|}{Text-to-Image}
    & \multicolumn{1}{c}{\multirow{2}{*}{RSUM}} \\
    \multicolumn{1}{c|}{}
    % &
    & R@1 & R@5 & R@10 & R@1 & R@5 & R@10 &
    & R@1 & R@5 & R@10 & R@1 & R@5 & R@10 & \\
\midrule %
\multicolumn{15}{l}{\textit{\textbf{ResNet-152 + Bi-GRU}}} \\ \midrule
VSE++~$_{\text{(BMVC' 18)}}$ \cite{faghri2017vse++}
    & 64.6 & 90.0 & 95.7 & 52.0 & 84.3 & 92.0 & 478.6
    & 41.3 & 71.1 & 81.2 & 30.3 & 59.4 & 72.4 & 355.7\\
PVSE~$_{\text{(CVPR' 19)}}$ \cite{song2019polysemous}
    & 69.2 & 91.6 & 96.6 & 55.2 & 86.5 & 93.7 & 492.8
    & 45.2 & 74.3 & 84.5 & 32.4 & 63.0 & 75.0 & 374.4\\
PCME~$_{\text{(CVPR' 21)}}$ \cite{chun2021probabilistic}
    & 68.8 & 91.6 & 96.7 & 54.6 & 86.3 & 93.8 & 491.8 & 44.2 & 73.8 & 83.6 & 31.9 & 62.1 & 74.5 & 370.1 \\
P2RM~$_{\text{(MM' 22)}}$ \cite{wang2022point}
    & 66.6 & - & - & 54.2 & - & - & -
    & 42.1 & - & - & 31.5 & - & - & -\\ 
DiVE~$_{\text{(CVPR' 23)}}$ \cite{kim2023improving}
    & 70.3 & 91.5 & 96.3 & 56.0 & 85.8 & 93.3 & 493.2
    & 47.2 & 74.8 & 84.1 & 33.8 & 63.1 & 74.7 & 377.7\\
\ccol \textbf{DVSE}
   & \ccol 69.1 & \ccol 92.5 & \ccol 96.8 & \ccol 55.7 & \ccol 86.7 & \ccol 93.5 & \ccol 494.2
   & \ccol 45.3 & \ccol 75.2 & \ccol 85.0 & \ccol 33.4 & \ccol 63.6 & \ccol 75.7 & \ccol 378.3 \\
\ccol \textbf{DVSE + FR}
   & \ccol \textbf{76.4} & \ccol \textbf{94.5} & \ccol \textbf{97.6} & \ccol \textbf{58.6} & \ccol \textbf{88.0} & \ccol \textbf{94.3} & \ccol \textbf{509.4}
   & \ccol \textbf{51.8} & \ccol \textbf{79.3} & \ccol \textbf{87.7} & \ccol \textbf{36.8} & \ccol \textbf{66.4} & \ccol \textbf{77.7} & \ccol \textbf{399.8} \\

\midrule %
\multicolumn{15}{l}{\textit{\textbf{Faster R-CNN + Bi-GRU}}} \\ \midrule

SCAN~$_{\text{(ECCV' 18)}}$ \cite{lee2018stacked}
    & 72.7& 94.8& 98.4& 58.8& 88.4& 94.8& 507.9
    & 50.4& 82.2& 90.0& 38.6& 69.3& 80.4& 410.9\\
MTFN-RR~$_{\text{(MM' 19)}}$ \cite{wang2019matching}
   & 74.3 & 94.9 & 97.9 & 60.1 & 89.1 & 95.0 & 511.3
   & 48.3 & 77.6 & 87.3 & 35.9 & 66.1 & 76.1 & 391.3\\
VSRN~$_{\text{(ICCV' 19)}}$ \cite{li2019visual}
    & 76.2 & 94.8 & 98.2 & 62.8 & 89.7 & 95.1 & 516.8
    & 53.0 & 81.1 & 89.4 & 40.5 & 70.6 & 81.1 & 415.7 \\
IMRAM~$_{\text{(CVPR' 20)}}$ \cite{chen2020imram}
    & 76.7 & 95.6 & 98.5 & 61.7 & 89.1 & 95.0 & 516.6
    & 53.7 & 83.2 & 91.0 & 39.7 & 69.1 & 79.8 & 416.5\\
VSE$_\infty$~$_{\text{(CVPR' 21)}}$ \cite{chen2021learning}
    & 78.5 & 96.0 & 98.7 & 61.7 & 90.3 & 95.6 & 520.8 
    & 56.6 & 83.6 & 91.4 & 39.3 & 69.9 & 81.1 & 421.9\\
SGRAF~$_{\text{(AAAI' 21)}}$ \cite{diao2021similarity}
   & 79.6 & 96.2 & 98.5 & 63.2 & 90.7 & 96.1 & 524.3
   & 57.8 & 91.6 & - & 41.9 & 81.3 & - & -\\
MVVSE~$_{\text{(IJCAI' 22)}}$ \cite{li2022multi}
    & 78.7 & 95.7 & 98.7 & 62.7 & 90.4 & 95.7 & 521.9
    & 56.7 & 84.1 & 91.4 & 40.3 & 70.6 & 81.6 & 424.6 \\
NAAF~$_{\text{(CVPR' 22)}}$ \cite{zhang2022negative}
   & 80.5 & 96.5 & 98.8 & 64.1 & 90.7 & 96.5 & 527.2
   & 58.9 & 85.2 & 92.0 & 42.5 & 70.9 & 81.4 & 430.9\\

DAA~$_{\text{(NeurIPS' 22)}}$ \cite{li2022differentiable}
    & 80.2 & - & - & 65.0 & - & - & -
    & 60.0 & - & - & 43.5 & - & - & -\\ 
DiVE~$_{\text{(CVPR' 23)}}$ \cite{kim2023improving}
    & 79.8 & 96.2 & 98.6 & 63.6 & 90.7 & 95.7 & 524.6 
    & 58.8 & 84.9 & 91.5 & 41.1 & 72.0 & 82.4 & 430.7 \\

% LG~$_{\text{(TMM' 23)}}$ \cite{li2023integrating}
%    & 79.6 & 96.5 & 98.5 & 64.4 & 90.9 & 95.9 & 525.8
%    & 59.1 & 84.9 & 92.0 & 43.0 & 71.8 & 82.2 & 432.9\\

\ccol \textbf{DVSE} 
   & \ccol 79.5 & \ccol 95.9 & \ccol 98.4 & \ccol 63.3 & \ccol 90.3 & \ccol 95.6 & \ccol 523.0
   & \ccol 58.8 & \ccol 85.2 & \ccol 91.7 & \ccol 41.2 & \ccol 71.0 & \ccol 81.5 & \ccol 429.3 \\
   
\ccol \textbf{DVSE + FR}
   & \ccol \textbf{84.5} & \ccol \textbf{97.0} & \ccol \textbf{98.8} & \ccol \textbf{66.1} & \ccol \textbf{91.6} & \ccol \textbf{96.2} & \ccol \textbf{534.1}
   & \ccol \textbf{65.1} & \ccol \textbf{87.7} & \ccol \textbf{92.9} & \ccol \textbf{44.9} & \ccol \textbf{73.9} & \ccol \textbf{83.6} & \ccol \textbf{448.2} \\

\midrule %
\multicolumn{15}{l}{\textit{\textbf{ResNeXt-101 + BERT}}} \\ \midrule
VSE$_\infty$~$_{\text{(CVPR' 21)}}$ \cite{chen2021learning}
    & 84.5 & 98.1 & 99.4 & 72.0 & 93.9 & 97.5 & 545.4 
    & 66.4 & 89.3 & 94.6 & 51.6 & 79.3 & 87.6 & 468.9  \\
DiVE~$_{\text{(CVPR' 23)}}$ \cite{kim2023improving}
    & 85.6 & 98.0 & 99.4 & 73.1 & 94.3 & 97.7 & 548.1 
    & 68.1 & 90.2 & 95.2 & 52.7 & 80.2 & 88.3 & 474.8 \\
\ccol \textbf{DVSE}
   & \ccol 84.8 & \ccol 97.8 & \ccol 99.4 & \ccol 70.9 & \ccol 93.3 & \ccol 97.1 & \ccol 543.4
   & \ccol 65.9 & \ccol 89.5 & \ccol 94.8 & \ccol 50.7 & \ccol 78.4 & \ccol 86.9 & \ccol 466.3 \\
   
\ccol \textbf{DVSE + FR} 
   & \ccol \textbf{90.2} & \ccol \textbf{98.4} & \ccol \textbf{99.5} & \ccol \textbf{72.1} & \ccol \textbf{93.5} & \ccol \textbf{96.9} & \ccol \textbf{550.7}
   & \ccol \textbf{74.1} & \ccol \textbf{92.1} & \ccol \textbf{96.0} & \ccol \textbf{53.4} & \ccol \textbf{79.9} & \ccol \textbf{87.5} & \ccol \textbf{483.0} \\
\bottomrule %
\end{tabular}
% }

% 
% \vspace{-5mm}
\label{tab:coco_comparison} 
\end{table*}

%% file: table/comparison_result_about_flickr30k.tex
\begin{table}[ht]
\centering
\caption{
The performance comparison between our model and other approaches on the Flickr30K dataset. We present the Recall@K and RSUM metric results. The best results are highlighted in bold.
}
% \scalebox{1}{
\footnotesize
\setlength{\tabcolsep}{2pt}
\begin{tabular}{l|ccc|ccc|c}
\toprule
    \multicolumn{1}{l|}{\multirow{2}{*}[0em]{Method}} 
    % & \multicolumn{7}{c|}{1K Test Images} 
    % & \multicolumn{7}{c}{5K Test Images} \\ \midrule
    % \multicolumn{1}{c|}{}
    % & 
    & \multicolumn{3}{c|}{Image-to-Text} 
    & \multicolumn{3}{c|}{Text-to-Image}
    & \multicolumn{1}{c}{\multirow{2}{*}{\textsc{RSUM}}}\\
    % \multicolumn{1}{c|}{}
    % &
    & R@1 & R@5 & R@10 & R@1 & R@5 & R@10 \\
\midrule %
\multicolumn{8}{l}{\textit{\textbf{ResNet-152 + Bi-GRU}}} \\ \midrule
VSE++~$_{\text{(BMVC' 18)}}$ \cite{faghri2017vse++}
    & 52.9 & 80.5 & 87.2 & 39.6 & 70.1 & 79.5 & 409.8\\
PVSE~$_{\text{(CVPR' 19)}}$ \cite{song2019polysemous} 
    & 59.1 & 84.5 & 91.0 & 43.4 & 73.1 & 81.5 & 432.6\\
PCME~$_{\text{(CVPR' 21)}}$ \cite{chun2021probabilistic} 
    & 58.5 & 81.4 & 89.3 & 44.3 & 72.7 & 81.9 & 428.1\\
DiVE~ $_{\text{(CVPR' 23)}}$ \cite{kim2023improving}
    & 61.8 & 85.5 & 91.1 & 46.1 & 74.8 & 83.3 & 442.6\\
\ccol \textbf{DVSE} 
   & \ccol 62.6 & \ccol 86.1 & \ccol 92.7 & \ccol 46.4 & \ccol 76.3 & \ccol 84.7 & \ccol 448.7\\
\ccol \textbf{DVSE + FR}
   & \ccol \textbf{70.0} & \ccol \textbf{88.8} & \ccol \textbf{93.4} & \ccol \textbf{51.6} & \ccol \textbf{78.8} & \ccol \textbf{86.7} & \ccol \textbf{469.4}\\

\midrule %
\multicolumn{8}{l}{\textit{\textbf{Faster R-CNN + Bi-GRU}}} \\ \midrule

SCAN~$_{\text{(ECCV' 18)}}$ \cite{lee2018stacked}
    & 67.4 & 90.3 & 95.8 & 48.6 & 77.7 & 85.2 & 465.0\\
MTFN-RR~$_{\text{(MM' 19)}}$ \cite{wang2019matching}
    & 65.3 & 88.3 & 93.3 & 52.0 & 80.1 & 86.1 & 465.1\\
VSRN~$_{\text{(ICCV' 19)}}$ \cite{li2019visual}
    & 71.3 & 90.6 & 96.0 & 54.7 & 81.8 & 88.2 & 482.6\\
IMRAM~$_{\text{(CVPR' 20)}}$ \cite{chen2020imram}
    & 74.1 & 93.0 & 96.6 & 53.9 & 79.4 & 87.2 & 484.2\\
VSE$_\infty$~$_{\text{(CVPR' 21)}}$ \cite{chen2021learning}
    & 76.5 & 94.2 & 97.7 & 56.4 & 83.4 & 89.9 & 498.1\\
SGRAF~$_{\text{(AAAI' 21)}}$ \cite{diao2021similarity}
    & 77.8 & 94.1 & 97.4 & 58.5 & 83.0 & 88.8 & 499.6\\
MVVSE~$_{\text{(IJCAI' 22)}}$ \cite{li2022multi}
    & 79.0 & 94.9 & 97.7 & 59.1 & 84.6 & 90.6 & 505.8\\
DAA~$_{\text{(NeurIPS' 22)}}$ \cite{li2022differentiable}
    & 78.0 & - & - & 59.9 & - & - & -\\ 
NAAF~$_{\text{(CVPR' 22)}}$ \cite{zhang2022negative}
    & 81.9 & 96.1 & 98.3 & 61.0 & 85.3 & 90.6 & 513.2\\
DiVE~$_{\text{(CVPR' 23)}}$ \cite{kim2023improving}
    & 77.8 & 94.0 & 97.5 & 57.5 & 84.0 & 90.0 & 500.8\\
% LG~$_{\text{(TMM' 23)}}$ \cite{li2023integrating}
%     & 81.5 & 95.9 & 98.3 & 61.0 & 85.1 & 90.4 & 512.2\\

\ccol \textbf{DVSE} 
   & \ccol 78.9 & \ccol 94.3 & \ccol 96.9 & \ccol 58.7 & \ccol 84.4 & \ccol 90.7 & \ccol 503.9\\
   
\ccol \textbf{DVSE + FR}
   & \ccol \textbf{85.6} & \ccol \textbf{96.2} & \ccol \textbf{98.0} & \ccol \textbf{64.5} & \ccol \textbf{87.5} & \ccol \textbf{92.7} & \ccol \textbf{524.5}\\

\midrule %
\multicolumn{8}{l}{\textit{\textbf{ResNeXt-101 + BERT}}} \\ \midrule
VSE$_\infty$~$_{\text{(CVPR' 21)}}$ \cite{chen2021learning}
    & 88.4 & 98.3 & 99.5 & 74.2 & 93.7 & 96.8 & 550.9\\
DiVE~$_{\text{(CVPR' 23)}}$ \cite{kim2023improving}
    & 88.8 & 98.5 & 99.6 & 74.3 & 94.0 & 96.7 & 551.9\\
\ccol \textbf{DVSE}
   & \ccol 88.4 & \ccol 98.9 & \ccol 99.5 & \ccol 75.0 & \ccol 93.7 & \ccol 96.7 & \ccol 552.1\\
   
\ccol \textbf{DVSE + FR} 
   & \ccol \textbf{94.9} & \ccol \textbf{99.4} & \ccol \textbf{99.9} & \ccol \textbf{77.1} & \ccol \textbf{94.2} & \ccol \textbf{96.9} & \ccol \textbf{562.4}\\
\bottomrule %
\end{tabular}
% }

% \vspace{-5mm}
\label{tab:flickr_comparison} 
\end{table}

%% file: sec/4_experiment.tex
\section{Experiments}
In this section, we conduct experiments with different feature encoders to validate the effectiveness of our approach. Firstly, we provide a comprehensive comparison between our method and state-of-the-art approaches. Subsequently, we validate each component of the model through different combinations of ablation experiments. We then perform sensitivity analyzes on the hyperparameters, and qualitatively analyze the retrieval results through visualization.

\subsection{Experimental Setup} 
\subsubsection{Datasets}
Experiments are conducted on three standard datasets \textbf{MSCOCO}, \textbf{Flickr30K} and \textbf{CUB Captions}. The MSCOCO and Flickr30K are sourced from different real-life scenarios, used to demonstrate model performance under heterogeneous data distribution. 
However MSCOCO and Flickr30K have a large number of false-negative samples \cite{chun2021probabilistic}, which are not conducive to evaluating model performance when one-to-many matching are made. Therefore, Fllowing \cite{chun2021probabilistic, wang2022point}, we chose CUB Captions as an additional evaluation dataset, which consists entirely of fine-grained bird categories, can be used to demonstrate model performance under homogeneous data distribution. Since image-text pairs belonging to the same category are considered as positive pairs, this dataset greatly suppresses false negative. We follow the data splits as defined in \cite{karpathy2015deep,frome2013devise,xian2017zero} for the three datasets. 

\subsubsection{Evaluation Metrics}
 For Flickr30K and MSCOCO, we use Recall@K (R@K) with K = 1, 5, 10 and RSUM as task evaluation metrics \cite{faghri2017vse++}. R@K measures the percentage of test samples where the correct match appears in the top K retrieved results, and RSUM is the sum of R@K for K = 1, 5, 10. For CUB Captions, we use R-Precision (R-P) for performance evaluation \cite{chun2021probabilistic}. This metric considers the order of rankings, and achieves the highest precision when and only when all positive samples are ranked before negative samples, which fully takes into account the relevant samples and thus better evaluates one-to-many matching results.
\subsubsection{Implementation Details}
To validate the effectiveness of our model and make a comprehensive comparison with previous methods, we perform experiments using various visual and textual encoders. For MSCOCO and Flickr30K, we employ three different backbones: ResNet-101 of Faster-RCNN \cite{ren2015faster} pre-trained on ImageNet and Visual Genome (BUTD) \cite{anderson2018bottom}, ResNeXT-101(32×8d) \cite{xie2017aggregated} pre-trained on Instagram (WSL) \cite{mahajan2018exploring}, and ResNet-152 \cite{he2016deep} pre-trained on ImageNet \cite{deng2009imagenet}. For the text encoder, we utilize both BiGRU \cite{cho2014learning} with Glove \cite{pennington2014glove} and BERT-Base \cite{devlin2018bert}. The embedding dimension ($D$) for each model is set to 1024. For CUB Captions, we use a ResNet-50 backbone and BiGRU with an embedding dimension of D = 512. To make a fair comparison, the image resolution is set to $512 \times 512$ for ResNeXT-101, and $224 \times 224$ for all other models. We utilize the AdamW optimizer with an initial learning rate of 5e-4. For the BUTD encoder, we trained for a total of 25 epochs, reducing the learning rate to 10\% of the original rate at the 15th epoch. The ResNet-50, ResNet-152 and ResNeXt-101 encoder models are trained for 30, 50, 50 epochs, with a learning rate reduction to 10\% at the 15th and 25th epochs. We set the hyperparameters $\alpha = 0.2$, $\beta = 0.4$. For CUB Captions, we set $K = 8$, $\gamma1 = 9$, $\gamma2 = 8$, $\lambda1 = 8$, $\lambda2 = 17$. For both MSCOCO and Flickr30K, we set $K = 6$, $\gamma1 = \gamma2 = 25$, $\lambda1 = \lambda2 = 20$. In ablation study, sensitive analysis and qualitative experiments, we use the Fast R-CNN + Bi-GRU as the feature encoders.

% 我们使用CUB数据集验证模型在one-to-many匹配上的精确性。能够避免假阴性
% PMRP主要检测语义丰富度，而Recall和RP都可以测试精度，但RP更加准确。
\input{table/comparison_result_about_CUB}
\input{table/FR_time}

\subsection{Comparisons with Baselines}
\subsubsection{Results On Heterogeneous Datasets}
TABLE \ref{tab:coco_comparison} and TABLE \ref{tab:flickr_comparison} present the performance of the compared methods on heterogeneous datasets MSCOCO and Flickr30K. We can see that DVSE outperforms the baseline model PVSE across different datasets and multiple metrics. For Flickr30K, DVSE improves the RUSM metrics by 3.7\%. This verifies the effectiveness of DVSE in optimizing set-based models. Without using FR, we achieve competitive or better results compared to previous set-based models \cite{li2022multi,song2019polysemous,kim2023improving}. We outperform the DiVE on Flickr30K in terms of RSUM. Meanwhile, compared to the BUTD encoder using an object detector to emphasize salient regions, ResNet-152 encoder which depicts the entire image has the better performance than DiVE on MSCOCO, indicating DVSE's ability to understand the semantics of the whole image and capture semantic variations. The combination of ResNeXt-101 and BERT we slightly inferior to VSE$\infty$ and DiVE on MSCOCO, but still exhibits similar performance. Unlike the designing of model structure of VSE$\infty$ and DiVE, DVSE emphasizes the improvement of losses, so their performances can complement each other. Additionally, FR improves the model performance significantly, surpassing other models on all metrics. This confirms the effectiveness of FR in resisting noise. Furthermore, since it operates directly on the similarity matrix, FR has the potential to enhance the accuracy of any image-text retrieval model. 

\subsubsection{Results on Homogeneity Dataset}
From the TABLE \ref{tab:cub_comparison}, it can be seen that as the set cardinality $K$ increases, the R@1 metric of PVSE keeps improving, but the R-P is decreasing. This indicates that PVSE focuses only on the most similar samples and does not handle the one-to-many matching relation well. On the contrary, DVSE obtains better R-P accuracy while showing similar R@1 accuracy, which indicates the effectiveness of DVSE in encouraging different sub-embeddings to encode different semantic variations. In addition, DVSE outperforms PCME and P2RM in R@1 metrics as a whole, which verifies the superiority of sub-embeddings in reducing information entropy. However, our model lags slightly behind the PCME and P2RM models on R-P, as these models typically emphasize retrieval diversity. After applying FR, all metrics show improvement, demonstrating the potential of FR in enhancing retrieval diversity and accuracy.

\subsubsection{Time Efficiency of Fast Re-ranking}
TABLE \ref{tab:time_comparison} shows the runtime and RSUM of our FR and other two re-ranking strategies on the Flickr30K and MSCOCO 5k test set. Compared to RR and MR, our method uses 0.4s in re-ranking on Flickr30K and is 7.8 and 33.8 times faster, and improves the accuracy by 2.0 and 2.2 times upon the baseline (i.e., w/o re-ranking), respectively. This observation validates the efficiency and effectiveness of FR. The computational complexity of the three re-ranking strategies are all \textbf{O(n)}, while FR maintains lower runtime and exhibits promising performance on MSCOCO 5K test set. So, FR is scalable and would show a stronger time advantage as the matrix size increases.

\begin{figure*}[tbp]
    \includegraphics[width=1.0\textwidth]{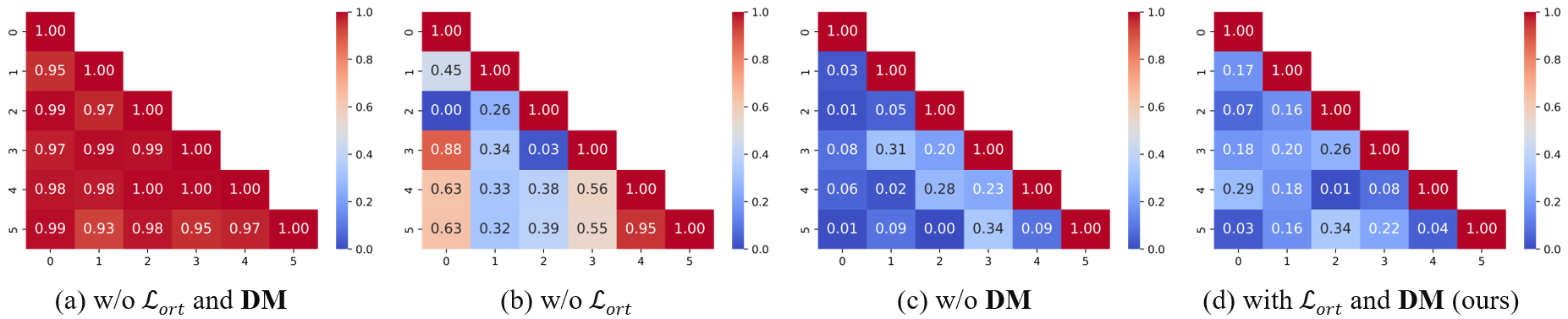}
    \caption{Visualization of correlation among visual sub-embeddings. We randomly select an image in MSCOCO and calculate the correlation coefficient matrix among its K sub-embeddings. Where the horizontal and vertical coordinates are the different sub-embeddings. Different figures show the ablation results of the ${L}_{ort}$ and the \textbf{DM}.}
\label{fig: correlation}
\end{figure*}

\input{table/ablation_result}

\input{table/ablation_views}

\begin{figure}[t]
\includegraphics[width=1.0\linewidth]{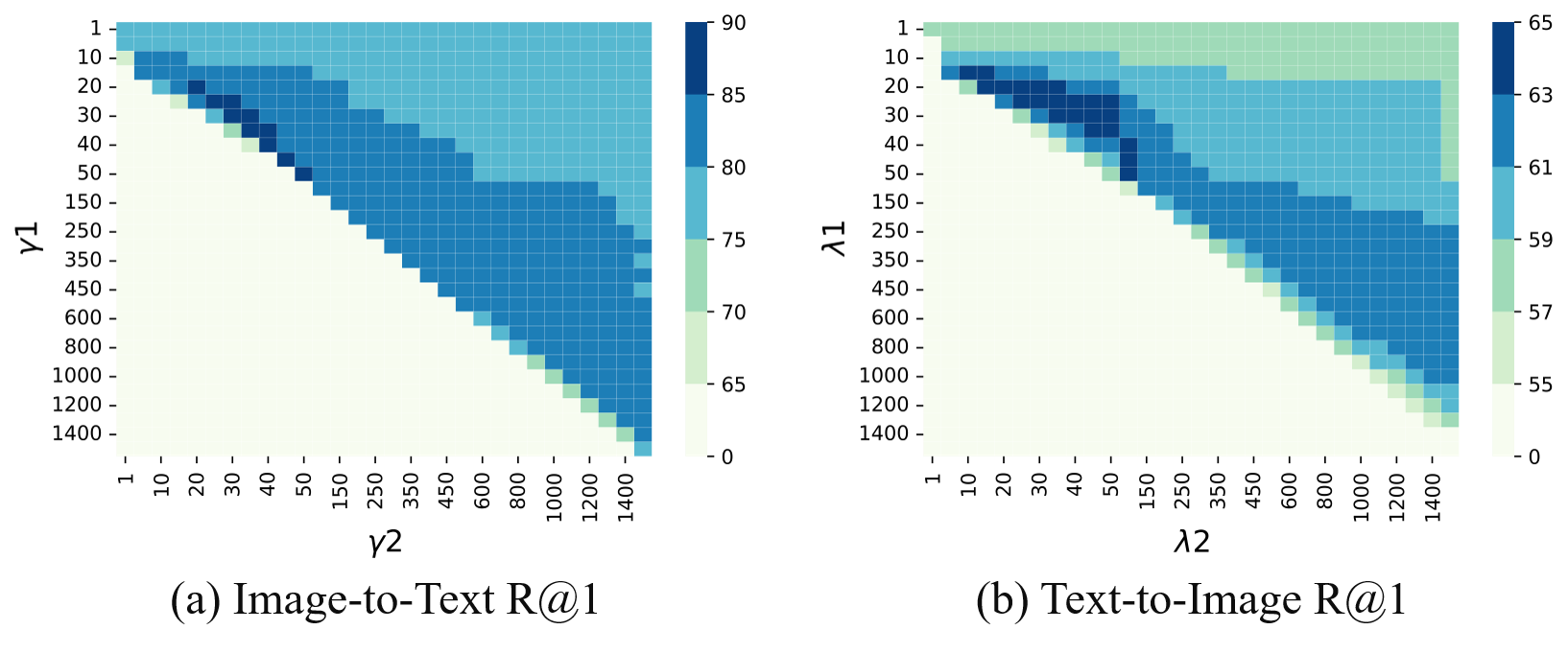}
\caption{The impact of scale parameters $\gamma1, \gamma2, \lambda1, \lambda2$. (a) and (b) show the R@1 results for image-to-text (row-wise retrieval) and text-to-image (column-wise retrieval) on Flickr30K, respectively. The horizontal and vertical axes are taken at intervals of 5 for values less than 50, 50 for values less than 500, and 100 for values less than 1500.}
\label{fig: scale}
\end{figure}

\begin{figure}[t]
\includegraphics[width=1.0\linewidth]{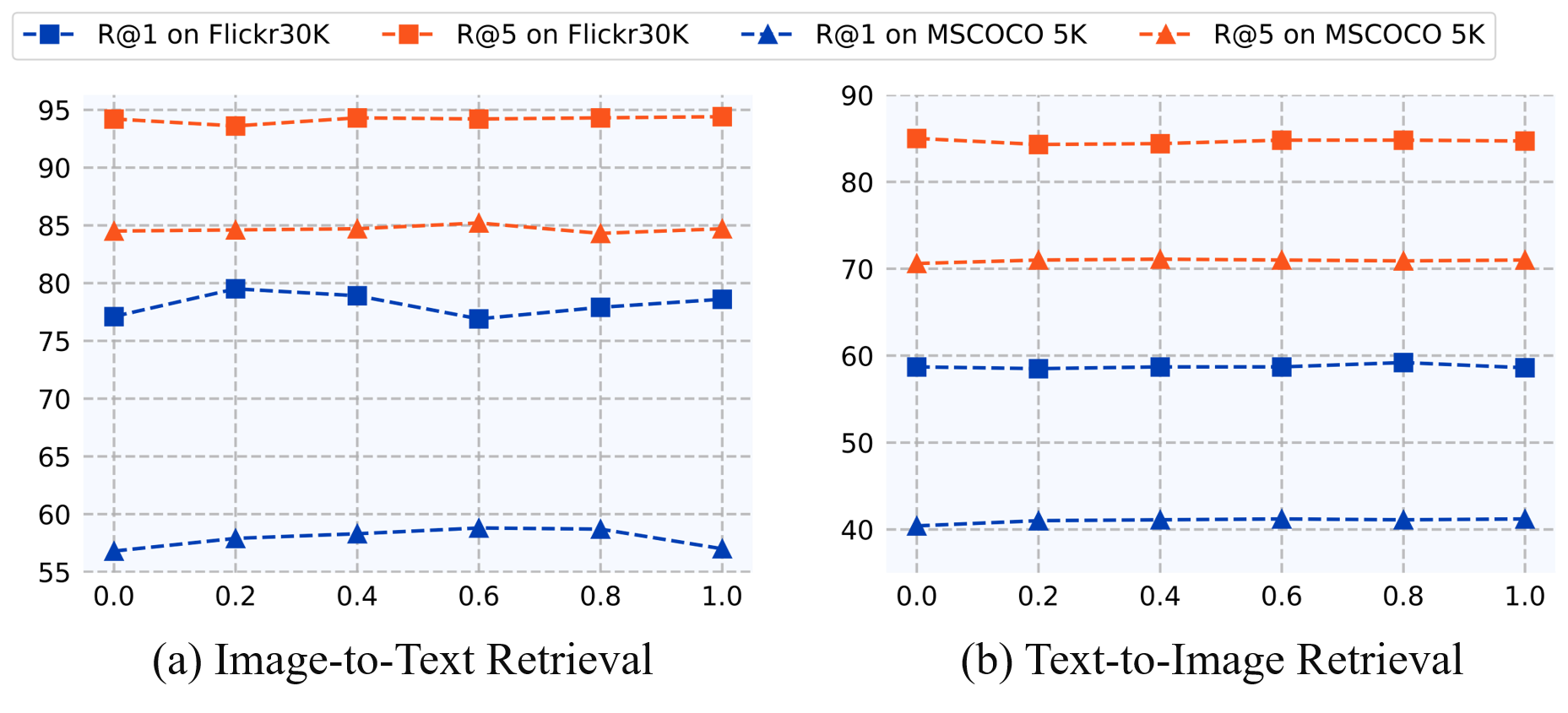}
\caption{The impact of the boundary parameter $\beta$, we show results for R@1 and R@5 on Flickr30K and MSCOCO 5K.}
\label{fig: boundary}
\end{figure}

\begin{figure}[t]
\includegraphics[width=1.0\linewidth]{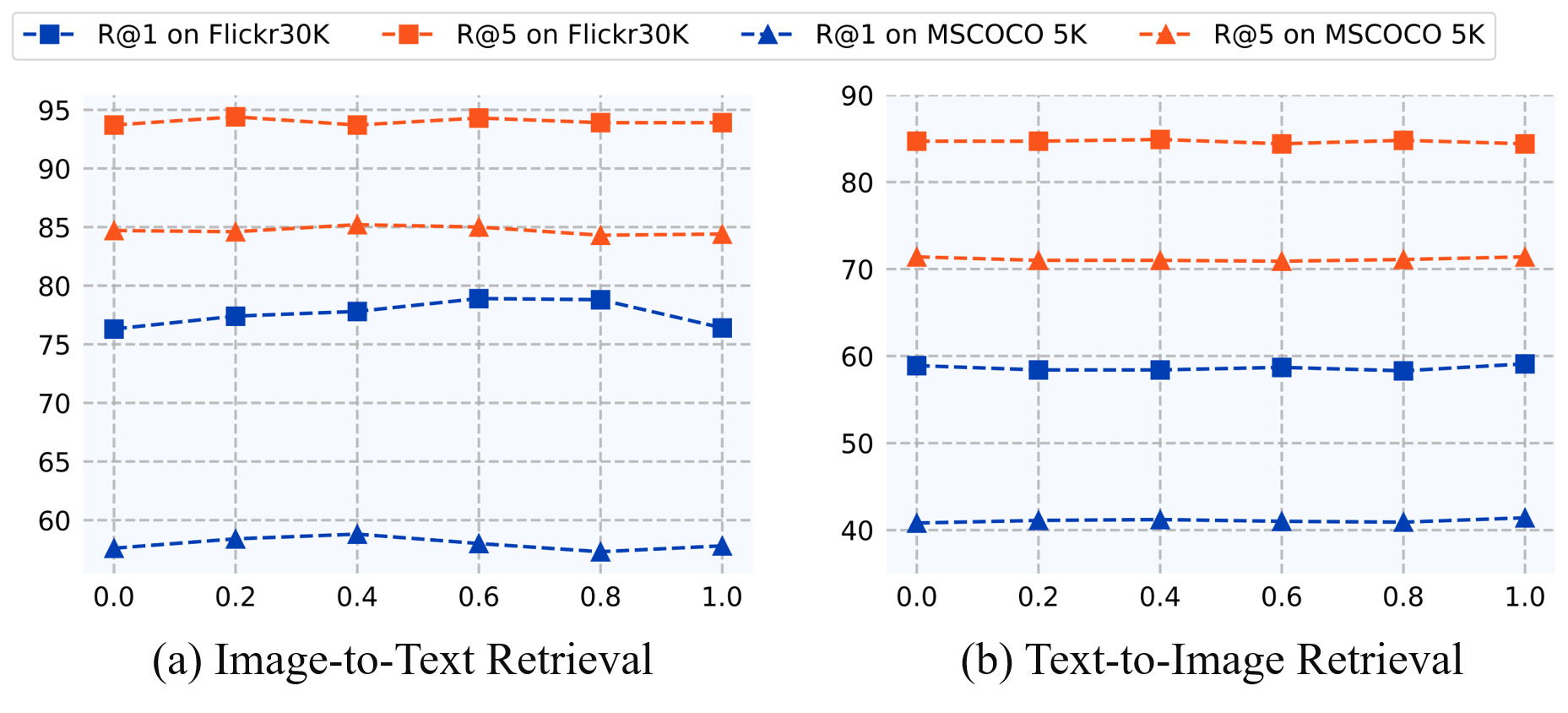}
\caption{The impact of the balancing factor $\eta$, we show results for R@1 and R@5 on Flickr30K and MSCOCO 5K.}
\label{fig: balance}
\end{figure}

\begin{figure*}[ht]
\includegraphics[width=1.0\linewidth]{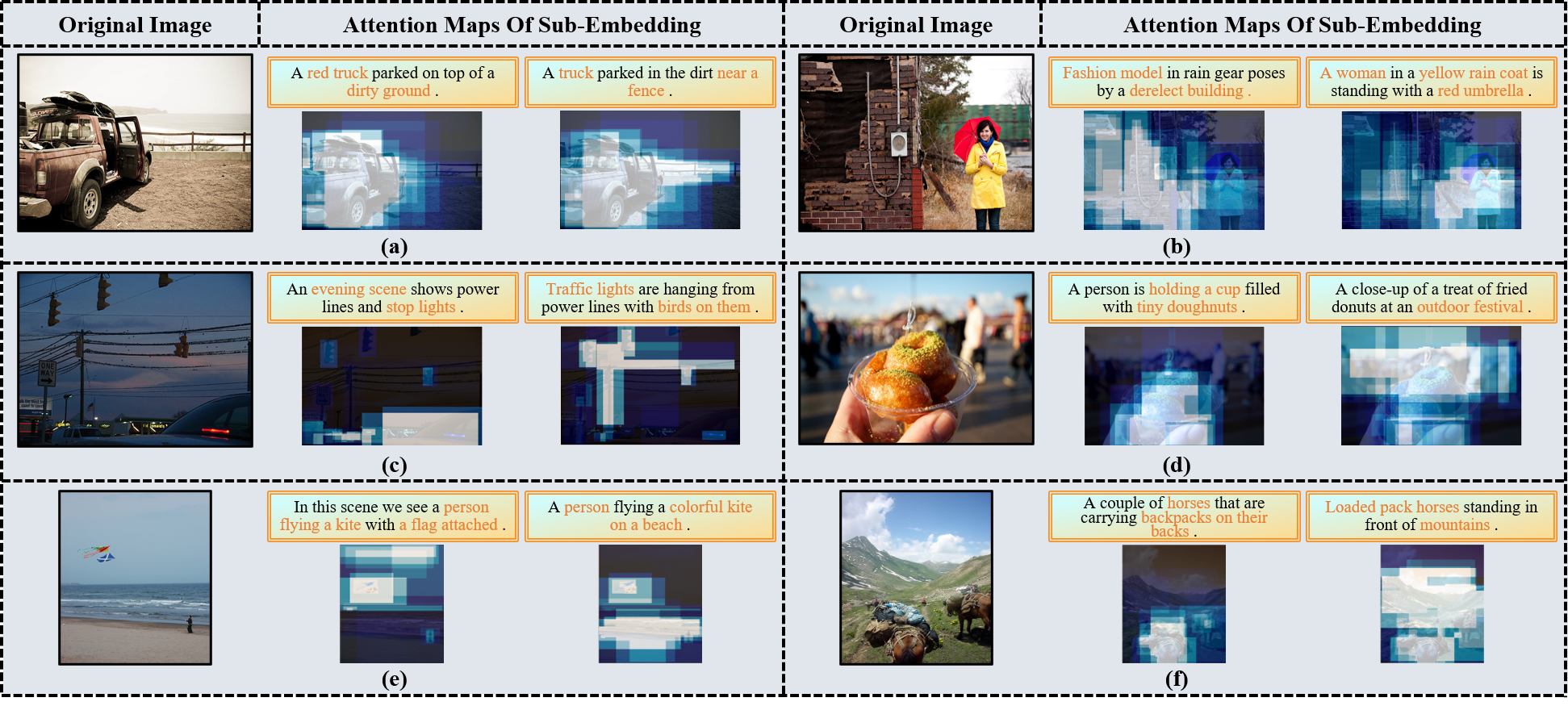}
\caption{Visualization of Sub-Embeddings. We evaluate the retrieval results on the MSCOCO dataset and present six distinct scenarios. Each scenario illustrates the original image, the corresponding textual description, and the associated heat maps.}
\label{fig: heatmap}
\end{figure*}
\begin{figure}[h]
\includegraphics[width=1.0\linewidth]{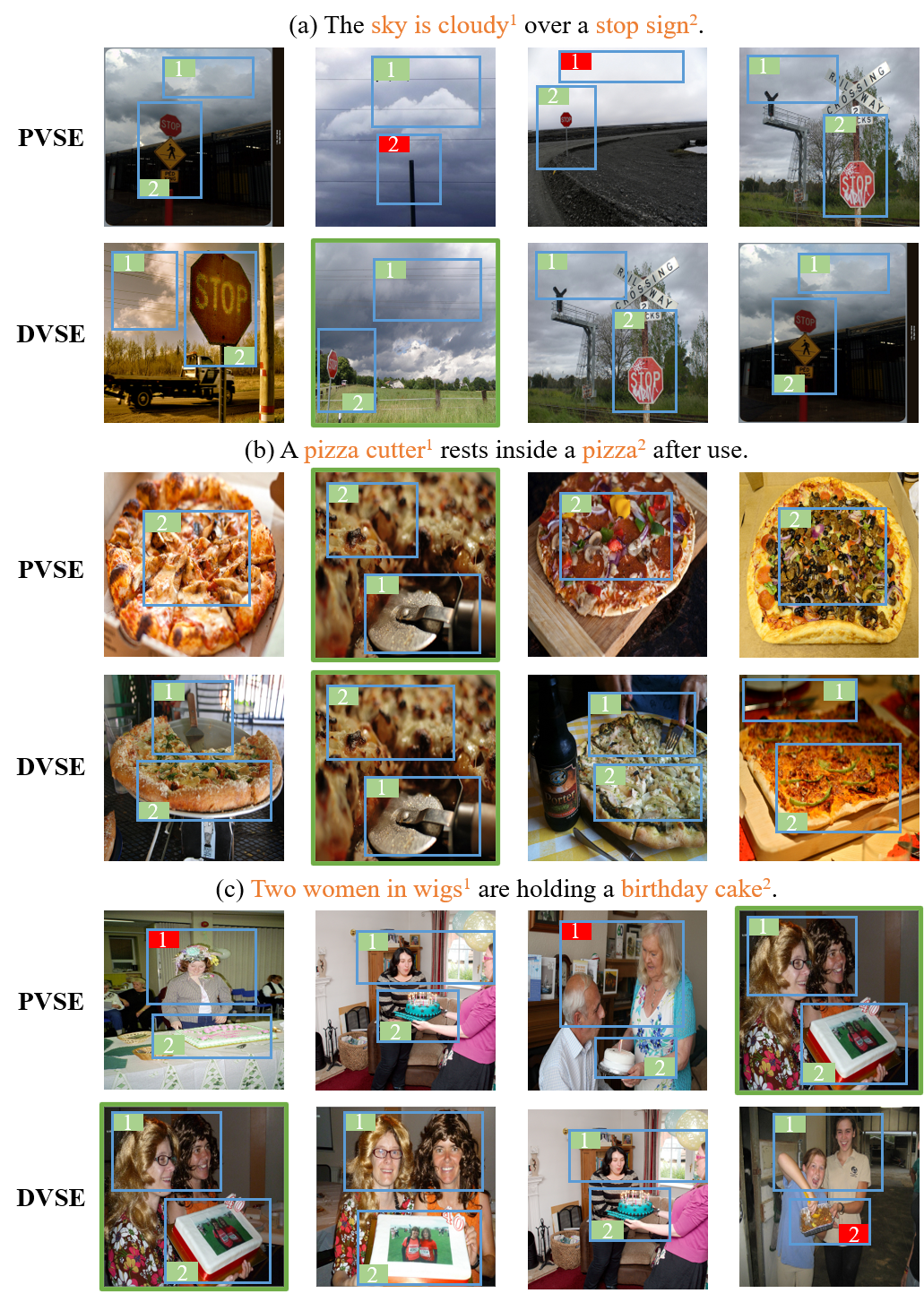}
\caption{Visualization of text-to-image retrieval results. We show the top 4 images retrieved from each text query, with the positive samples labeled in the dataset boxed in green. We use orange color and ordinals to highlight details in the text, and mark the ordinals of the corresponding regions and non-corresponding regions in the images in green and red, respectively.}
\label{fig: retrieval_ti}
\end{figure}
\begin{figure}[h]
\includegraphics[width=1\linewidth]{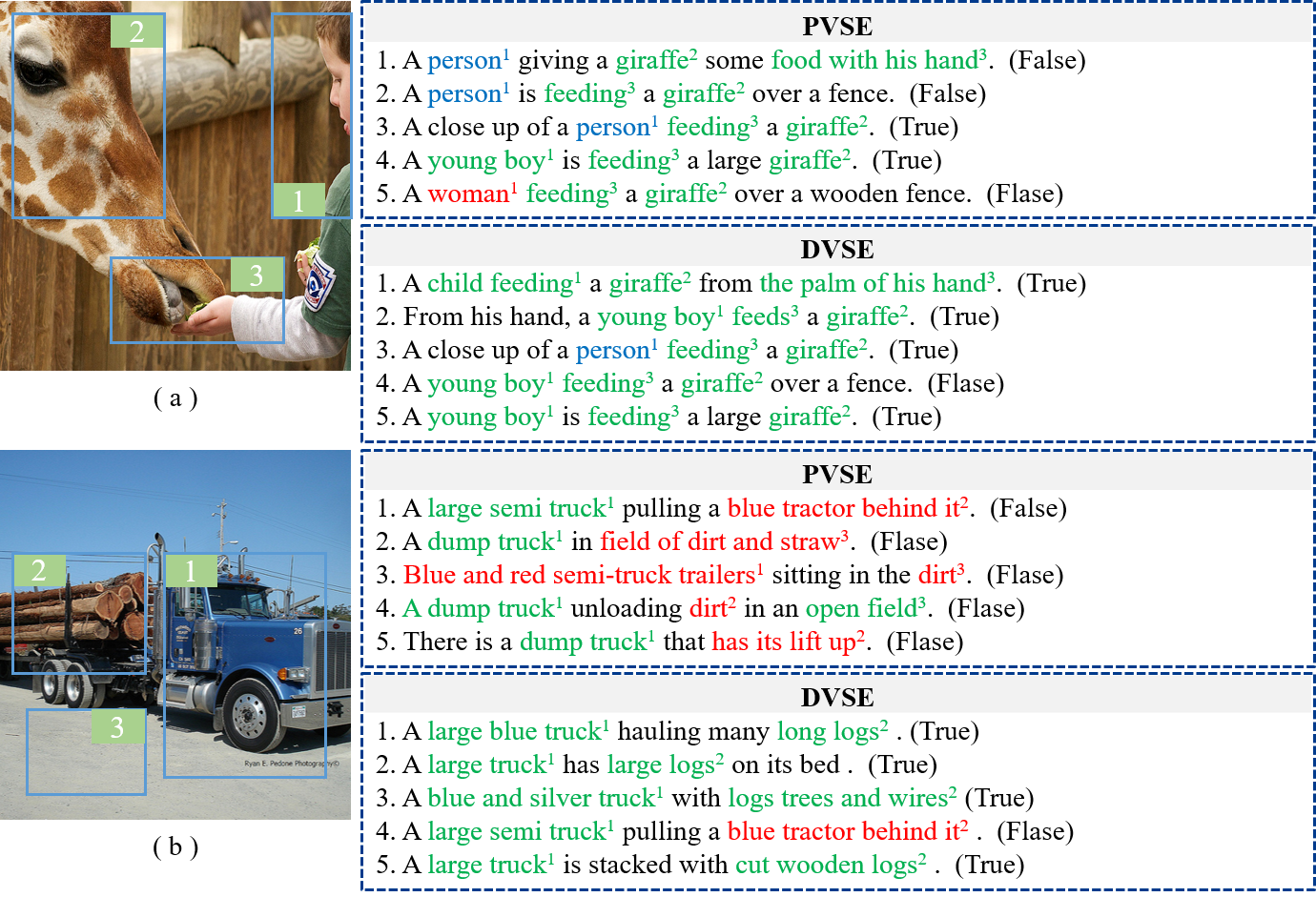}
\caption{Visualization of Image-to-Text Retrieval Results. We present the top 5 textual descriptions retrieved from two images respectively, with positive samples labeled in the dataset noted as `True' at the end of the sentence. We frame the details in the images with ordinals and mark corresponding words in the text in green, ambiguous words in blue, and errors in red.}
\label{fig: retrieval_it}
\end{figure}

\subsection{Ablation Studies}
TABLE \ref{tab:elements} and Fig. \ref{fig: correlation} show the ablation study on Flickr30K, which contains the dynamic orthogonal constraint loss ($\mathcal{L}_{dc}$), the variance-aware weighting loss ($\mathcal{L}_{var}$), and the sub-embedding (\textbf{SE}). For $\mathcal{L}_{dc}$, we decomposed it into two components to explore their roles respectively: orthogonal loss $\mathcal{L}_{ort}$ and dynamic masking \textbf{DM}, where $\mathcal{L}_{ort} = \left[\sum_{i=1}^{K}\sum_{j=1,j \neq i} ^{K}||(\hat{\mathbf{v}}^{i})^{T}\hat{\mathbf{v}}^{j}||-\beta\right]_{+}$. 
The experiment results indicate that $\mathcal{L}_{var}$ has the most significant impact on model performance, followed by $\mathcal{L}_{ort}$, \textbf{DM} and \textbf{SE}. The ablation combinations verified that DVSE can reduce information entropy and prevent set collapse.
\begin{itemize} 
\item Comparisons in Combination 4 and Combination 5 show that the use of \textbf{SE} alone does not improve model performance. Combined with the higher correlation exhibited between the sub-embeddings in Fig. \ref{fig: correlation}(a), it can be concluded that in the absence of effective loss supervision, different embeddings learn the same information and thus fall into set collapse. This validates the effectiveness of DVSE in optimizing set-based models and preventing set collapse.

\item Combination 3 ablates $\mathcal{L}_{var}$ and \textbf{DM}, and it can be observed that when the sub-embeddings are not optimized by the weights, the model performance decreases to a level close to the full ablation. Thus only reducing the correlation between embeddings is not sufficient to motivate the model to learn semantic variations, and the embeddings are still likely to degrade to a high entropy state. This validates the role of $\mathcal{L}_{var}$ in reducing the entropy of embeddings.

% \item The comparison between Combination 3 and Combination 7 demonstrates that even in the absence of $\mathcal{L}_{var}$, \textbf{DM} still effectively enhances the model's performance. This indicates that dynamic masking can enhance the optimization ability of the model for all sub-embeddings and avoid local optimal.

% \item The comparison between Combination 4 and Combination 5 reveals that solely using sub-embeddings also does not alter the accuracy. This is because when loss supervision is lacking, different embeddings learn the same information. This validates the importance of the combination of the two proposed losses in guiding sub-embeddings to learn the semantic variation.

\item The strong ability of ${L}_{ort}$ to reduce the correlation of sub-embeddings is presented in Fig. \ref{fig: correlation}(c), while \textbf{DM} prevents the model from falling into the local optimum. As shown in Fig. \ref{fig: correlation}(b), without directly constraining orthogonality, \textbf{DM} still reduces the correlation between sub-embeddings. It indicates that the model is able to focus on optimization of multiple sub-embeddings, thus encouraging them to learn different semantic information. Furthermore, compared to Fig. \ref{fig: correlation}(d), we find that the correlation in the non-ablation model increases with the improvement of accuracy (table \ref{tab:elements}), which indicates that the DVSE is not simply decomposing the semantics, but learns the semantic relations from the data.
\end{itemize}

% \subsubsection{Correlation Between Sub-Embeddings} \label{S: CBS}
% Fig. \ref{fig: correlation} illustrates the impact of the proposed ${L}_{ort}$ and \textbf{DM} on the correlation between image sub-embeddings. As depicted in (a) and (d), there is a high correlation between the sub-embeddings when both orthogonal loss and DM are absent, which validates the important role of $L_{dc}$ in assisting the model to encode semantic variations. \text{DM} prevent the model from falling into the local optimum. As shown in (b), with the assistance of \text{DM}, the correlation of sub-embeddings is weakened and the model is able to focus on the training of multiple sub-embeddings. The strong ability of ${L}_{ort}$ to reduce the correlation of sub-embeddings is presented in (c). Compared to (d), we find that the correlation in the non-ablation model increases with the improvement of accuracy (table \ref{1}), which indicates that the model is not simply decomposing the semantics, but learns the semantic relations from the data.

% As depicted in (a) and (d), 当正交损失和DM同时被消融时，子嵌入间存在较高的相关性，这验证了$\matchal{L}_{dc}$在辅助模型编码语义变化的重要作用。

%\text{DM}可以保证每个子嵌入都能得到训练进而防止模型陷入局部最优。如图(b)所示，在\text{DM}的辅助下，子嵌入的相关性得到了一定的减弱，模型能够关注多个子嵌入的训练。

% 图c展示的结果中反映了${L}_{ort}$降低子嵌入相关性的强大能力。当与图d进行对比时，我们发现完整结果中嵌入间的相关性有一定的提升，这说明模型并非简单地分解语义，而是能够从数据中学习到相应的语义关联。

%本文的主要目的是解决基于集的模型所出现的两种崩塌现象，从图中是可以看出，
% 1、正交损失能够显著降低子嵌入间的相关性。
% 2、方差损失也能够在一定程度上降低相关性。
% 3、方差损失能够提高正交损失所降低的一部分相关性。
% 第一点自然是说明了正交损失的有效性，但其所带来的未必是好的结果，因为不同语义变化之间确实存在着一定的联系。

% 语义变化应该是对同一张图片包含的不同语义。

\subsection{Sensitivity Analysis}
In this section, we analyze the influence of key parameters on accuracy, including set cardinality $K$ in Equation \eqref{eq: loss_var}, the boundary parameter $\beta$ in Equation \eqref{eq: loss_ort}, the balancing factor $\eta$ in Equation \eqref{eq: loss_final} and the scale parameters $\gamma1, \gamma2, \lambda1, \lambda2$ in Equation \eqref{eq: normal param}.
\subsubsection{Cardinality Of Sub-Embedding Set}
The cardinality of the sub-embedding set $K$ reflects the extent to which the model deconstructs semantic variations. We set $\beta$ and $\eta$ to 0.4 and 0.6 respectively and examined the impact of different cardinalities on the CUB Captions and MSCOCO 5K datasets. To exclude the influence of FR on the results, we ablate it in this experiment. TABLE \ref{tab:views} shows that there is a significant increase in model performance when using SE compared to without SE. For MSCOCO, after $K = 4$, the increasing trend of accuracy slows down, and the best accuracy is achieved at $K = 6$, followed by a slight decrease in accuracy. It demonstrates that an appropriate set cardinality can effectively capture semantic variations, while too high a set cardinality may introduce noise redundancy. The results on CUB Captions show the same pattern, but the optimal accuracy occurs at a larger set cardinality, i.e., $K = 8$. This observation reveals that heterogeneous data usually contain easily identifiable features, whereas homogeneous data are more difficult to distinguish and require a larger set cardinality to identify semantic variations.

\subsubsection{Scale Parameters}
We use $\gamma1$, $\gamma2$, $\lambda1$, $\lambda2$ to adjust the tightness of \eqref{eq: normalize}, and when $\gamma1 < \gamma2$ or $\lambda1 < \lambda2$, the contribution of the elements in the original matrix is emphasized. The row-wise retrieval in Fig. \ref{fig: scale}(a) shows that when $\gamma1$ and $\gamma2$ are equal and in the interval from 1 to 20, R@1 keeps increasing. In the interval from 20 to 50, R@1 reaches the optimum and then starts to decrease when they are greater than 50. That is because FR cannot introduce effective discriminative information if $\gamma1$ and $\gamma2$ are too low; while the similarity distribution is destroyed when they are too high. Meanwhile, we note that when $\gamma1 > \gamma2$, there is a significant decrease in R@1. That means the contribution of the elements in the original matrix is weakened, and the information from the columns is not sufficient to rank the results. A decrease in performance is also observed for $\gamma1 < \gamma2$, which demonstrates the effectiveness of the discriminative information from column introduced by the FR. The column-wise retrieval in \ref{fig: scale} (b) exhibits the same pattern; however, the position of its optimal value in the matrix is shifted upward from diagonal. This is because an image in the dataset corresponds to more than one text, so to avoid confusing other positive samples with negative samples, the contribution of the elements in the original matrix needs to be emphasized.

% TABLE \ref{tab:views} demonstrates that a lower cardinality significantly reduces performance. This is due to the fact that  As the number of $K$ increases, the accuracy of the model increases, suggesting that sub-embeddings can capture semantic variations. (在不使用SE 时，精度有较为显著的差异，当基数开始增加，模型精度有了明显的提升，当集合基数达到4以后，模型精度开始稳定，当基数为2时，the number of sub-embeddings is not enough to encode semantic variations, and the loss unsuccessfully supervise the optimization process.  而较高的模型基数则能够成功捕捉语义变化)
% Nevertheless, higher $K$ may introduce noise into the model, leading to instability results.

\subsubsection{Boundary Parameter and Balancing Factor}
We also analyze the influence of $\beta$ and $\eta$ on Flickr30K and MSCOCO 5K test. As shown in Fig. \ref{fig: boundary}, $\beta$ has a stronger impact on image-to-text retrieval, while it is stable on text-to-image retrieval. This is because the 1:5 ratio between image and text leads to more sensitive results in that direction. The R@1 on the two datasets peaks at $\beta=0.2$ and $0.6$, respectively, and then presents the trend of decline and fluctuation. The results for parameter $\eta$ are similar to that of $\beta$ and shows fluctuations on image-to-text retrieval in Fig. \ref{fig: balance}. The optimal R@1 on the two datasets is obtained at $\eta=0.6$ and $0.4$, respectively. $\beta$ is used to adjust the correlation boundary for $\mathcal{L}_{dc}$, while $1-\eta$ emphasizes the contribution of $\mathcal{L}_{dc}$ in the objective function, so the R@1 in image-to-text retrieval on Fig. \ref{fig: boundary} and Fig. \ref{fig: balance} has a tendency to complement each other. In general, the two parameters are sensitive to model performance in image-to-text retrieval and need to be set carefully. Based on above observations, we fix $\beta=0.4$ and $\eta=0.6$ when comparing to the baseline for fairness.
% 我们同样探索了
% (思考该实验的描述位置)We also explore the hyperparameters of Fast Re-ranking on retrieval accuracy and diversity. （详细说明为什么这样简化）For clarity, we simplify the Fast Re-ranking strategy by assuming $\gamma_{1} = \gamma_{2}$ and $\lambda_{1} = \lambda_{2}$. As shown in Fig. \ref{fig: diversity and accuracy}, with varying hyperparameters, both PMRP and R@1 are significantly improved compared to the results of ablation Fast Re-ranking, highlighting the effectiveness of the Fast Re-ranking strategy. Furthermore, PMRP experiences a rapid decrease before R@1 reaches its peak, indicating that Fast Re-ranking may sacrifice some diversity to further enhance accuracy.

\subsection{Qualitative Results}
In this section, we use heat maps and bidirectional retrieval results to qualitatively verify the ability of DVSE to capture semantic variations. For a fair comparison, we did not utilize FR here.
\subsubsection{Heat Map of Sub-Embeddings}
To intuitively demonstrate the image semantics captured by sub-embeddings, we use the heat maps [16, 17]. Specifically, for each sub-embedding vector, we computed its similarity with $R$ image region vectors extracted by Faster R-CNN. Subsequently, based on the magnitude of similarity, we arranged all regions in descending order and assigned attention scores according to their position index $r_{i}$ as $a_{i} = \rho \times (R-r_{i})^{2}$, where $\rho$ is a coefficient. Finally, the attention score for each pixel was calculated by aggregating the scores of all associated regions. As depicted in Fig. \ref{fig: heatmap}, we selected various typical scenarios, including individuals, landscapes, animals, transportation, etc. The results demonstrate that different sub-embeddings effectively capture semantic variations in the images corresponding to the provided text. For instance, in (c), (d) and (e), when the text describes the images from different perspectives, the regions of interest also change accordingly. In addition, we note that the model does not simply decompose the image, as this may result in isolated semantics. As shown in (a), (b), and (f), when the text semantics are expanded, the model's regions of interest are also expanded to new regions on top of the original ones, which in turn emphasizes the semantic connections. Overall, the regions of interest of DVSE are able to track the relevant semantics and make corresponding changes and additions, proving its effectiveness and robustness.

\subsubsection{Visualization of Retrieval Results}
Fig. \ref{fig: retrieval_ti} and Fig. \ref{fig: retrieval_it} illustrate the top 4 retrieval results of DVSE and PVSE in bidirectional retrieval. As shown in Fig. \ref{fig: retrieval_ti}, although the retrieved images are all relevant to textual semantics, compared to PVSE, DVSE can recognize more detailed information. In Fig. \ref{fig: retrieval_ti}(a), PVSE confuses the difference between the stop sign and the chimney. In Fig. \ref{fig: retrieval_ti}(b), the text contains the object `pizza cutter'. Although PVSE retrieves one positive sample, this information is missing from the other three images. Similarly, in Fig. \ref{fig: retrieval_ti}(c), DVSE captures key factors such as the number and gender of people, and retrieve four images containing these details, although the cake in the fourth image is not correctly matched. Fig. \ref{fig: retrieval_it} also demonstrates better noise resilience in DVSE. As depicted in Fig. \ref{fig: retrieval_it}(a), the DVSE use `boy' and `child' to describe the age of the character while PVSE makes more use of the ambiguous pronoun `person'. In addition, PVSE in Fig. \ref{fig: retrieval_it}(b) does not recognize the cargo carried by the truck, suffers from interference from the environment, such as `dirt' and `straw'. In contrast, although DVSE also makes an error in the fourth result, the results are more focused on detailed regions such as `track' and `logs'.

%% file: table/comparison_result_about_CUB.tex
\begin{table}[t]
\centering
\caption{
The performance comparison between our model and other approaches on the CUB dataset. We present the Recall@1 and R-P metric results. The best results are highlighted in bold.
}
% \scalebox{1}{
\footnotesize
\setlength{\tabcolsep}{3pt}
\begin{tabular}{l|cc|cc}
\toprule
    \multicolumn{1}{l|}{\multirow{2}{*}[0em]{Method}} 
    % & \multicolumn{1}{c|}{\multirow{3}{*}[-1.6em]{CA}} 
    % & \multicolumn{4}{c}{1K Test Images} \\ \midrule
    
    & \multicolumn{2}{c}{Image-to-Text} 
    & \multicolumn{2}{c}{Text-to-Image}\\
    & R-P & R@1 & R-P & R@1 \\
\midrule 
VSE++ $_{\text{(BMVC' 18)}}$ \cite{faghri2017vse++} & 22.4 & 44.2 & 22.6 & 32.7\\
\midrule %
PVSE K=1 $_{\text{(CVPR' 19)}}$ \cite{song2019polysemous} & 22.3 & 40.9 & 20.5 & 31.7\\
PVSE K=2 $_{\text{(CVPR' 19)}}$ \cite{song2019polysemous}& 19.7 & 47.3 & 21.2 & 28.0\\
PVSE K=4 $_{\text{(CVPR' 19)}}$ \cite{song2019polysemous}& 18.4 & 47.8 & 19.9 & 34.4\\
\midrule %
PCME $_{\text{(CVPR' 21)}}$ \cite{chun2021probabilistic} & 26.3 & 46.9 & 26.8 & 35.2\\
P2RM $_{\text{(MM' 22)}}$ \cite{wang2022point} & 26.8 &47.1 & \textbf{27.9} & 37.2\\

\ccol \textbf{DVSE}
   & \ccol 26.2 & \ccol 47.4 & \ccol 26.6 & \ccol 37.1\\
\ccol \textbf{DVSE + FR} 
   & \ccol \textbf{27.4} & \ccol \textbf{50.3} & \ccol 26.8 & \ccol \textbf{37.2}\\

\bottomrule %
\end{tabular}
% }

% \vspace{-5mm}
\label{tab:cub_comparison} 
\end{table}

%% file: table/FR_time.tex
\begin{table}[ht]
\centering
\caption{
The computational efficiency between our model and other approaches on the Flickr30K dataset. We present the Test Time and RSUM metric results. The best results are highlighted in bold.
}
% \scalebox{1}{
\footnotesize
\setlength{\tabcolsep}{10pt}
\begin{tabular}{l|cc}
\toprule
    \multicolumn{1}{l|}{Method} 
    % & \multicolumn{7}{c|}{1K Test Images} 
    % & \multicolumn{7}{c}{5K Test Images} \\ \midrule
    % \multicolumn{1}{c|}{}
    % & 

    % \multicolumn{1}{c|}{}
    % &
    & \textsc{Test Time (s)} & \textsc{RSUM}\\
\midrule %
\multicolumn{3}{l}{\multirow{1}{*}[0em]{Flickr30K}}\\
\midrule %
DVSE
     & 1.5 & 503.9\\
DVSE + RR $_{\text{(MM' 19)}}$ \cite{wang2019matching}
     & 4.6 & 514.0\\  
DVSE + MR $_{\text{(TGRS' 22)}}$ \cite{yuan2022remote}
     & 15.0 & 513.4\\
\ccol \textbf{DVSE + FR (Ours)}
    & \ccol \textbf{1.9} & \ccol \textbf{524.5}\\
\midrule %
\multicolumn{3}{l}{\multirow{1}{*}[0em]{MSCOCO}}\\
\midrule %
DVSE
     & 47.7 & 429.3\\
DVSE + RR $_{\text{(MM' 19)}}$ \cite{wang2019matching}
     & 113.5 & 436.7\\  
DVSE + MR $_{\text{(TGRS' 22)}}$ \cite{yuan2022remote}
     & 447.2 & 439.6\\
\ccol \textbf{DVSE + FR (Ours)}
    & \ccol \textbf{56.3} & \ccol \textbf{448.2}\\
\bottomrule %
\end{tabular}
% }
\vspace{-5mm}
\label{tab:time_comparison} 
\end{table}

%% file: table/ablation_result.tex
\begin{table}[t]
\centering
\caption{
Ablation Study of our model, including Dynamic Mask (\textbf{DM}), Variance-Aware Weighting Loss ($\mathcal{L}_{var}$), Orthogonal Loss ($\mathcal{L}_{ort}$), and Sub-Embeddings (\textbf{SE}).
}
% \scalebox{1}{
\footnotesize
\setlength{\tabcolsep}{6pt}
\begin{tabular}{c|cccc|c}
\toprule
    \multicolumn{1}{c|}{\multirow{2}{*}[0em]{Combination}} 
    & \multicolumn{4}{c|}{Ablation in Flickr30K}  \\
    & $\mathbf{DM}$ & $\mathcal{L}_{var}$ &$\mathcal{L}_{ort}$ & $\mathbf{SE}$ & RSUM \\ 
\midrule
    1 &\ccol\bcmark& \ccol\bcmark &\ccol\bcmark& \ccol\bcmark & \ccol\textbf{503.9} \\
    2 & & \bcmark & \bcmark & \bcmark & {502.8} \\
    3 & & & \bcmark & \bcmark & {500.6} \\
    4 & & & &\bcmark  & {499.8} \\
    5 & & & & & {499.8} \\ 
\midrule
    6 & \bcmark & \bcmark & & \bcmark & {501.9} \\ 
    7 & \bcmark & & \bcmark & \bcmark & {501.4} \\ 
\bottomrule
\end{tabular}
% }

% \vspace{-5mm}
\label{tab:elements}
\end{table}

%% file: table/ablation_views.tex
\begin{table}[t]
\centering
\caption{
The impact of the sub-embedding set cardinality on CUB Captions and MSCOCO 5K. The best results are highlighted in bold.
}
% \scalebox{1}{
\footnotesize
\setlength{\tabcolsep}{4pt}
\begin{tabular}{l|c|c|c|c|c|c|c}
\toprule
    \multicolumn{1}{l|}{\multirow{1}{*}[0em]{Method}} 
    % & \multicolumn{1}{c|}{\multirow{3}{*}[-1.6em]{CA}} 
    % & \multicolumn{4}{c}{1K Test Images} \\ \midrule
    
    % & \multicolumn{2}{c}{Image-to-Text} 
    % & \multicolumn{2}{c}{Text-to-Image}\\
    & w/o SE & K=2 & K=4 & K=6 & K=8 & K=10 & K=12\\
% \midrule 
% \multicolumn{3}{l}{\multirow{1}{*}[0em]{Flickr30K}}\\
% \midrule %
% Image-Text R@1 & 61.6 & 75.9 & 76.9 & \textbf{78.9} & 77.6 & 77.3 & 77.7\\
% Text-Image R@1 & 29.2 & 56.8 & 58.4 & \textbf{58.7} & 58.4 & 58.7 & 58.7\\
% RSUM & 408.1 & 496.8 & 500.2 & \textbf{503.9} & 501.7 & 502.3 & 502.1\\
\midrule 
\multicolumn{3}{l}{\multirow{1}{*}[0em]{CUB Captions}}\\
\midrule %
I-T R@1 & 34.5 & 36.3 & 42.2 & 46.9 & \textbf{47.4} & 45.8 & 47.1\\
I-T R-P & 21.3 & 21.0 & 24.9 & 25.8 & \textbf{26.2} & 25.7 & 26.4\\
T-I R@1 & 26.3 & 28.3 & 34.9 & 35.4 & \textbf{37.1} & 36.2 & 36.5\\
T-I R-P & 20.6 & 21.6 & 24.6 & 26.3 & \textbf{26.6} & 26.3 & 26.8\\

\midrule %
\multicolumn{3}{l}{\multirow{1}{*}[0em]{MSCOCO 5K}}\\
\midrule %
I-T R@1 & 37.1 & 55.7 & 57.1 & \textbf{58.8} & 58.0 & 57.3 & 57.2\\
T-I R@1 & 25.6 & 38.7 & 40.8 & \textbf{41.2} & 40.8 & 41.0 & 41.0\\
RSUM & 324.5 & 413.8 & 426.8 & \textbf{429.3} & 426.9 & 428.1 & 426.3\\
\bottomrule %
\end{tabular}
% }

% \vspace{-5mm}
\label{tab:views} 
\end{table}

%% file: sec/5_conclusion.tex
\section{Conclusion}
This paper explores the construction of embeddings in cross-modal retrieval from the perspective of information entropy, and addresses the challenges of set collapse encountered in set-based models. Specifically, we propose a Dynamic Visual Semantic Sub-Embeddings (DVSE) framework with the Fast Re-ranking (FR) strategy. This framework applies dynamic orthogonal constraint loss to encode the semantic variations in images, and motivate different sub-embeddings to capture different semantics through variance-aware weighting loss. FR further boosts the model's resilience to noise by incorporating a maximum discrimination term into the ranking process, and yields a higher time efficiency. The experiments on MSCOCO and Flickr30K datests with heterogeneous data distributions, as well as CUB Captions with homogeneous data distributions, have validated the effectiveness of our method. In future work, we will further reduce the embedding entropy while mining richer semantic variations with the help of language-image pre-training mechanism, to better model the semantic ambiguity and improve the retrieval performance.